%% file: main.tex
\documentclass{artificial-life}

\usepackage{color}
\usepackage[dvipsnames]{xcolor}

\usepackage{soul}      
\colorlet{r1}{white}
\colorlet{r2}{white}
\newcommand{\Hl}[2][\empty]{%
    \ifx#1\empty
    \else
    \sethlcolor{#1}%
    \fi
    \hl{#2}}
\newcommand{\citeg}[1]{\citep[e.g.,][]{#1}}
\soulregister\Hl{7}
\soulregister\citeauthor7
\soulregister\citeyearpar7
\soulregister\cite7
\soulregister\citep7
\soulregister\citet7
\soulregister\citeg7
\soulregister\ref7
\soulregister\pageref7
\soulregister\gls7
\soulregister\glspl7
\soulregister\footnote7
\soulregister\csquare7

\usepackage{caption}
\usepackage{subcaption}

\usepackage{amsthm}
\usepackage{amssymb}

\theoremstyle{definition}
\newtheorem{definition}{Definition}

\usepackage[nopostdot,style=super,nonumberlist,toc]{glossaries}
\setacronymstyle{long-short}
\input{glossary}
\usepackage{orcidlink}
\usepackage{ragged2e}

\usepackage{tikz}
\newcommand{\csquare}[1]{%
    \raisebox{-0.06cm}{%
        \tikz{%
            \ifthenelse{\equal{#1}{black}}{%
                \fill[gray!70] (0,0) rectangle (0.3,0.3); 
                \draw[black,step=0.1cm] (0,0) grid (0.3,0.3); 
            }{%
                \fill[gray!70] (0,0) rectangle (0.3,0.3); 
            }
        }%
    }%
}

\usepackage{fancyhdr}

\newcommand{\orcid}[1]{\href{orcid.org/#1}{\textcolor[HTML]{A6CE39}{\aiOrcid}}}

\usepackage[style=apa,natbib=true]{biblatex}

\begin{document}

\pagestyle{fancy}
\fancyhf{} 
\fancyfoot[L]{
\begin{minipage}{\textwidth}\ \\[12pt]\copyright\ 2025 Massachusetts Institute of Technology.
https://doi.org/10.1162/ARTL.a.10 
This is the author’s final version and has been accepted for publication in Artificial Life.
\end{minipage}
} 
\renewcommand{\headrulewidth}{0pt} 

\begin{titlepage}
    \begin{center}
        {\small \textit{Artificial Life} Manuscript Submission \par}
        {\large Untapped Potential in Self-Optimization of Hopfield Networks: The Creativity of Unsupervised Learning \par}
        \vspace{1cm}
        {Natalya Weber \affil{1} \orcidlink{0000-0002-1955-3612} ,
        Christian Guckelsberger \affil{2,3} \twitter{creativeendvs} \orcidlink{0000-0003-1977-1887},
        Tom Froese \affil{1} \twitter{DrTomFroese} \orcidlink{0000-0002-9899-5274} \par}
        {\textbf{Corresponding:} Natalya Weber (natalya.weber@oist.jp) \par}
        \vspace{0.5cm}
        {\footnotesize \begin{flushleft}
        \item 1. Embodied Cognitive Science Unit, Okinawa Institute of Science and Technology Graduate University, Okinawa, 904-0412, Japan
        \item 2. Department of Computer Science, Aalto University, 02150, Espoo, Finland
        \item 3. School of Electronic Engineering and Computer Science, Queen Mary, University of London, London, E1 4NS, United Kingdom \end{flushleft}
        \par}
        \vspace{1cm}
        {\justifying
        \textbf{Abstract.} The \gls{so} model can be considered as the third operational mode of the classical \acrlong{hn}, leveraging the power of associative memory to enhance optimization performance. Moreover, it has been argued to express characteristics of minimal agency, which renders it useful for the study of artificial life. In this article, we draw attention to another facet of the \gls{so} model: its capacity for creativity. 
        \Hl[r1]{Drawing on creativity studies, we argue that the model satisfies the necessary and sufficient conditions of a creative process. Moreover, we show that learning is needed to find creative outcomes above chance probability. Furthermore, we demonstrate that modifying the learning parameters in the \gls{so} model gives rise to four different regimes that can account for both creative products and inconclusive outcomes, thus providing a framework for studying and understanding the emergence of creative behaviors 
        in artificial systems that learn.} 
        \par}
        \vspace{1cm}
        {\begin{flushleft}
         \end{flushleft} \par}
        \vfill
    \end{center}

    {\small \copyright\ 2025 Massachusetts Institute of Technology. \\
    https://doi.org/10.1162/ARTL.a.10 \\
    This is the author’s final version and has been accepted for publication in Artificial Life. \par}
    \end{titlepage}



\section{Introduction \label{sec:Introduction}}
\glsreset{hn}
\glsreset{so}
Understanding the phenomenon of life, including its origins and potential for change \Hl[r1]{during or across lifetimes}, remains an open challenge for science. A fundamental methodological issue is the difficulty of distinguishing between life's essential and contingent features, given that on our planet we only have a sample size of 1 -- all extant living beings share a common ancestor. However, although a principled comparative study of life itself is impossible, synthetic approaches provide a practical alternative solution: we can explore the space of possibilities by studying life as it could be - \gls{alife} \citep{langton_preface_1989}. 

One important insight arising out of \gls{alife} is the importance of agency for the phenomenon of life \citep[e.g.][]{froese_behavior-based_2012,froese_motility_2014}. Life is inherently active: from the small scale of biochemistry to the ever-larger scales of behavior, learning, development, and reproduction, organismic activity is not a mere event or happening, it is a purposeful activity. And this means that what an organism does could potentially go wrong, too - not everything works equally well, and even an organism's very existence is at stake when it becomes trapped for too long in a suboptimal trajectory. 
And yet, despite its inherently precarious existence, life is surprisingly resilient \citep{ball_how_2025}. In this maintenance and enhancement of their viability, organisms must also meet novel situations with purposeful action.

A second essential insight coming from the field of \gls{alife} is that the phenomenon of life is characterized by the ability to continuously reinvent itself in an open-ended manner \citep{song_little_2022}; life is always capable of surprising us with novelty and - even more impressively - to do so in a highly context-sensitive and adaptive manner. Often, this innovation contributes to an organism's fitness, resonates with us aesthetically \citep{boden_creativity_2015}, or both. \Hl[r1]{Thus, even if not explicitly labeled as such in core definitions of the phenomenon of life \citeg{bedau_four_1998}, open-ended innovation often also has \emph{value}}.

\emph{Novelty} and \emph{value}, whether benefiting an organism's survival or our aesthetic pleasure, are both considered defining properties of creativity \citep{runco_ai_2023}; it is not a coincidence that the field of \gls{alife} has a long-standing interest in creativity and includes artistic explorations \citeg{penny_art_2009,dorin_artificial_2015,wu_survey_2024}. \Hl[r1]{
However, creativity researchers have mostly studied people, and given that the concept of creativity was originally used exclusively with reference to humans and the divine \citeg{still_history_2016}, its definitions that followed are anchored in our lived experiences as people, often relying on the presence of human cognitive faculties. Some studies have considered some highly developed animals, such as the family of Corvidae \citep{kaufman_animal_2015}, while less attention has been paid to other (potentially less complex) lifeforms, let alone life itself.
It seems plausible, nevertheless, to assume that if agency and creativity are characteristics of life, then they are not independent from each other. But what precisely could be the nature of their interdependence?} 

\Hl[r1]{The main challenge in answering this question lies in integrating insights across various fields of investigation and scales of description. Accordingly, the goal of this paper is to make a small step toward the development of a conceptual framing that could enable us to investigate the interdependence between agency and creativity in formal terms and simpler, life-like systems. Specifically, we will address the question: Does a simple model of agency also satisfy the conditions of creativity? To do so, we draw on advances in \gls{alife} that connect a mathematical model of a complex adaptive system \citep{watson_optimization_2011}, namely a variation of the famous Hopfield Network \citep{hopfield_neural_1982}, with theoretical considerations of agency \citep{watson_agency_2024}. As our core contribution, we investigate how the same model can be productively interpreted from the perspective of contemporary theories in creativity research. This alignment of interpretive frameworks points to the exciting possibility of a unified theory of life's agency and creativity, which future work could unfold more fully. We only briefly draw on the relationship of open-endedness and creativity. In \gls{alife}, \emph{open-endedness} is typically studied at an evolutionary timescale across generations of individuals. Thus, more future work is needed to connect our proposal on the relation between agency and creativity to those debates. 
}

\Hl[r1]{While our focus is on \gls{alife}, our work can also benefit the field of creativity research. Here, the need for a dynamical perspective on creativity that accounts for the mere potential to be creative receives increasing support \citep{corazza_potential_2016,beghetto_dynamic_2019,green_process_2023}. In Sec.~\ref{sec:SO-and-CC} we show that learning in the \gls{so} model results in four different regimes that can account for both creative products and inconclusive outcomes and ``failures'', thus providing a framework for studying the creative potential. Moreover, our analysis of the probability of creative products may provide a mathematical framework that can complement methods like that of \citet{simonton_defining_2018} for distinguishing between diverse kinds of uncreative ideas based on their probability and utility.}

\Hl[r1]{This work continues a young line of research which strives to better understand creativity in artificial systems (and, potentially, the natural systems they seek to model) by mapping between formal models from \gls{ai} research, and definitions as well as theories of creativity from creativity research  \citep{lahikainen_creativity_2024}. In contrast to previous work though, which operated on Markov Decision Processes as formalisations of sequential decision making \emph{problems}, this work operates on the \gls{so} model as a concrete and minimal formalism to \emph{solve} such problems.}  

\glsreset{hn}
\glsreset{so}
\section{Background \label{sec:Background}}

Many stories can be told about the 2024 Nobel Prize in Physics that was awarded to John J. Hopfield and Geoffrey E. Hinton \textquotedblleft \emph{for foundational discoveries and inventions that enable machine learning with artificial neural networks}\textquotedblright{} \citep{nobel_foundation_press_2024}. \textbf{\glspl{hn}} are mostly known for two operational modes: they can recall patterns as an associative memory \citep{hopfield_neural_1982,hopfield_neurons_1984-1}, or they can compute a solution to a constraint optimization problem \citep{hopfield_neural_1985}. \Hl[r1]{As we will see in Sec.~\ref{subsec:HN-dynamics}, the key difference between these two modes is that in the associative memory mode, the desired information is explicitly stored in the network, whereas in the constraint optimization mode, the network is used to implicitly define and compose the sought-after information.} It was not until more than 20 years after Hopfield's original work that \citet{watson_effect_2009} proposed another operational mode for \glspl{hn}: by leveraging the power of associative memory, the system can learn to optimize its behavior towards some desirable goal state encoded in the network. \Hl[r1]{By combining the two modes, the third mode uses the network to distill a single generalized piece of information from the specified properties. \citeauthor{watson_effect_2009} }termed this \Hl[r1]{mode} the \textquotedblleft self-modeling\textquotedblright{} framework \citep{watson_optimization_2011}, which we are going to address as the \gls{so} model\footnote{This change was made to sidestep unresolved debates about the existence and representational status of internal models.}  \citep{zarco_self-optimization_2018,froese_autopoiesis_2023}. We contribute an analysis of how this largely underexplored, additional operational mode of \glspl{hn}, can be used to model creativity. 

The computational modeling of creativity falls in the realm of \textbf{\gls{cc}} research \citep{colton_computational_2012}. Existing work falls on a continuum  \citep{perez_computational_2018-1} between the \emph{cognitive perspective}, for which researchers devise models to produce insights into the nature of creativity, and the \emph{engineering perspective}, denoting a concern for engineering \gls{ai} systems that exhibit some form of creativity, autonomously or in human-machine co-creation, but without necessarily simulating human cognition. Crucially though, most existing \gls{cc} research can be located at the engineering end of this continuum. Moreover, the few examples taking a cognitive perspective focus on \emph{human} creativity and cognitive \Hl[r2]{faculties} \citep{olteteanu_cognition_2020}. These insights shed little light on the relationship of creativity and agency, and it is unclear to \Hl[r2]{what} extent they apply to life more generally.

Initially, the \textbf{\glsreset{so}\gls{so}} model was mainly investigated in the context of abstract problems in the \gls{alife} community and applied to questions in theoretical biology \citep{watson_transformations_2011-3,morales_self-optimization_2019,gershenson_self-organization_2020}. \Hl[r2]{\citet{watson_optimization_2011} pointed out that the type of problems the \gls{so} model can potentially solve represent combinatorial problems, specifically Propositional Satisfiability problems, and recently, \cite{weber_use_2023-1} demonstrated this capability on several concrete examples}. Here, we continue this agenda by proposing that the \gls{so} model has a bilateral potential usage for research on creativity. On the one hand, employing the cognitive perspective, we aim to improve our understanding of creativity in livings beings. On the other hand, employing the engineering perspective, we consider the model's use for the design of artificial systems expressing creativity as it could be.


The rest of this article is structured as follows: in Sec.~\ref{sec:Classical-Hopfield-Networks} we describe the classical \glspl{hn} and their typical two operational modes \citep{hopfield_neural_1982,hopfield_neural_1985}, on which the \gls{so} model is based. In Sec.~\ref{sec:The-Self-Optimization-model}, we describe the \gls{so} model \citep{watson_optimization_2011}. In Sec.~\ref{sec:CC} we analyze how the \gls{so} model is informed by the product- and process-based perspectives on creativity. In Sec.~\ref{sec:SO-and-CC}, we evaluate how the \gls{so} model can be used for new insights in \gls{cc} research. In Sec.~\ref{sec:Learning-effort}, we discuss the relationship between learning and time constraints. In Sec.~\ref{sec:Discussion}, we discuss the findings and the implications of this work. Finally, in Sec.~\ref{sec:Conclusion}, we draw the conclusions of this work and discuss the various paths for future research.

\section{HNs \& Their Typical Uses\label{sec:Classical-Hopfield-Networks}}
\glsreset{hn}
In this Section we will introduce the key equations and resulting dynamics of \glspl{hn} together with intuitions. \Hl[r1]{We suggest readers with an existing understanding of \glspl{hn} and its two classical operational modes to skip straight to Sec.~\ref{sec:The-Self-Optimization-model} describing the \gls{so} model.} For a deeper coverage of the underlying physics and math, consider \citet{rojas_neural_1996} and \citet{hertz_introduction_2019}. \glspl{hn} can be broadly categorized into three types: discrete-time discrete-state \citep{hopfield_neural_1982}, discrete-time continuous-state \citep{koiran_dynamics_1994}, and continuous-time continuous-state \citep{hopfield_neurons_1984-1}. \Hl[r1]{We rest our argument on the discrete-time discrete-state type of \gls{so} models. However, the \gls{so} model was also shown to work for the discrete-time continuous-state \citep{zarco_self-modeling_2018} and continuous-time continuous-state \citep{zarco_self-optimization_2018}. We use the discrete-time discrete-state case as the arguably most accessible demonstration, but expect our conclusions to similarly hold for the continuous variants.}

\subsection{The Dynamics of HNs\label{subsec:HN-dynamics}}

A \gls{hn} is a representation of a physical system with $N$ nodes (or ``neurons''). It is a type of \Hl[r2]{\gls{rnn}}, where partial computations of the network are fed back to the network itself. In the case of the \gls{hn}, it is a fully connected \gls{rnn}, \Hl[r1]{i.e., each node is connected to every other node.}
The \Hl[r2]{state of the system} can be represented by a state vector $\mathbf{S}=\left\{s_{1},\ldots,s_{N}\right\} $, where a node $s_{i}$ can have binary values from $\left\{ 0,1\right\} $ or $\left\{ -1,1\right\} $ (also known as a \textit{bipolar} notation), or continuous values between $\left[0,1\right]$ or $\left[-1,1\right]$, respectively. The connections between the nodes are defined by a weight matrix, $\mathbf{W}$, of size $N\times N$, and the states are updated in asynchronous fashion\footnote{The states can be also updated in synchronous fashion; \citet[p.~3088]{hopfield_neurons_1984-1} introduced asynchrony deliberately ``to represent a combination of propagation delays, jitter, and noise in real neural systems.''} (i.e., at each time step, only one node $i$, chosen at random, is updated) according to the following rule for the bipolar case:
\begin{equation}
s_{i}\left(t+1\right)=f\left[\sum_{j}^{N}w_{ij}s_{j}\left(t\right)+I_{i}\right],\label{eq:state_update}
\end{equation}
where $w_{ij}$ are elements of the weight matrix $\mathbf{W}$, and $f$ is an activation function. $I_{i}$ allows to provide an external input to the node $i$ (e.g., sensory input) or to introduce an offset bias. For the discrete bipolar case, $f=\Theta$,
the Heaviside threshold function assumes values -1 for negative arguments and +1 otherwise (and similarly 0 or +1 for the binary case). Conversion of the \textbf{state update equation} (\ref{eq:state_update}) to and from binary notation of $q_{i}\in\left\{ 0,1\right\} \forall i$ can be done by substituting $s_{i}=2q_{i}-1$. 

\begin{figure}[t!]
     \centering
     \begin{subfigure}[b]{0.34\textwidth}
         \centering
         \includegraphics[width=\textwidth]{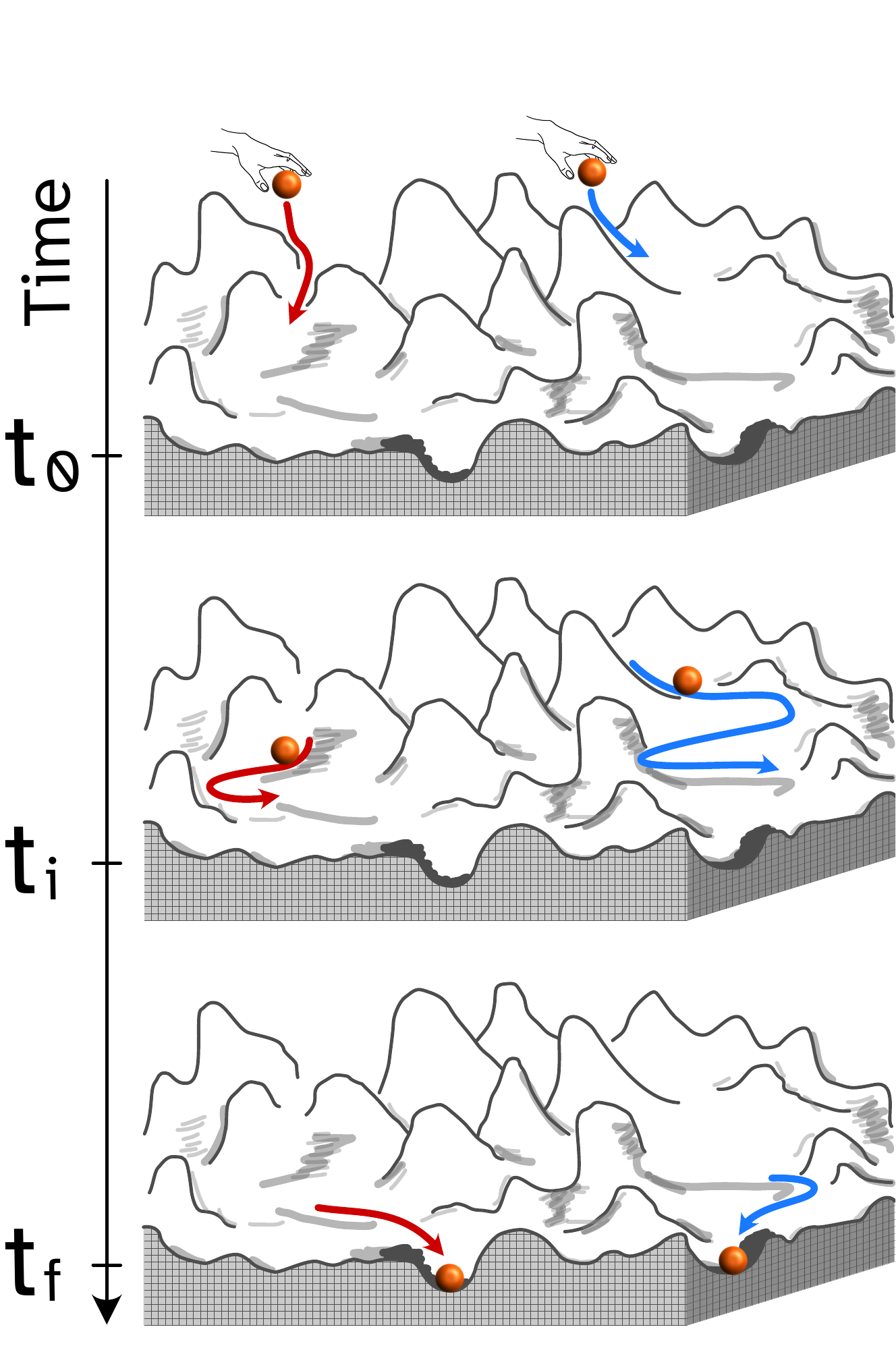}
         \caption{Classical two modes}
         \label{fig:classical-hns-dyn}
     \end{subfigure}
     \hspace{0.14cm}
     \begin{subfigure}[b]{0.6\textwidth}
         \centering
         \includegraphics[width=\textwidth]{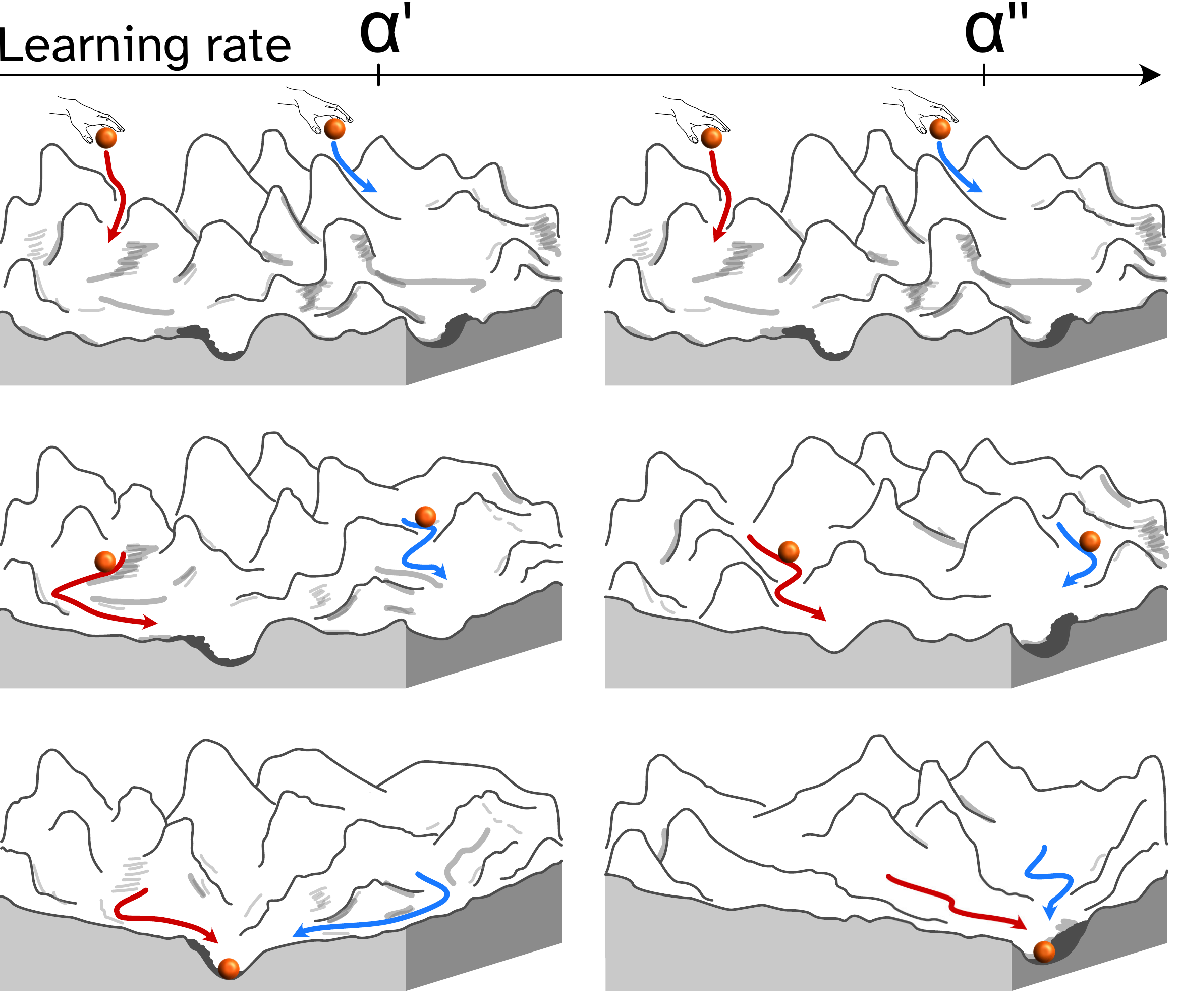}
         \caption{Third mode: self-optimization}
         \label{fig:so-dyn}
     \end{subfigure}
        \caption[Comparison of landscapes]{Dynamics of the three operational modes of \glspl{hn}, distinguished via the rigidness of the landscape metaphor. In all three modes, the initial landscape at $t_{0}$ is defined by the initial weight matrix $\mathbf{W}_{0}$, and the hands holding the two balls represent two possible initial states of the system (here visualized by the two horizontal degrees of freedom). (a) In the classical \gls{hn}, the weight matrix $\mathbf{W}_{0}$ does not change and as a result the landscape is static from $t_{0}$ to $t_{f}$. The minimum that the system will converge to highly depends on the initial conditions of the system. (b) In the \gls{so} model, the dynamics depend on the learned weight matrix $\mathbf{W}_{\mathrm{L}}$, that is constantly updated and as a consequence the landscape is constantly changing. With many consecutive Hopfield optimizations, the energy landscape is adjusted to deepen and widen the minima such that at some point, $t_{f}$, only one global attractor is left (second column). Regardless of the initial conditions, the system will converge to that state. Crucially, depending on the specifics of the initial weight matrix and the choice of learning rate $\alpha$, the final global attractor may or may not be the initial global minimum (third column). \Hl[r1]{In the figure, ground shading \csquare{black} indicates a static landscape, while \csquare{gray} represents a changing landscape.}}
        \label{fig:The-landscapes}
\end{figure}

We can analyze the difference between system states by computing the \textbf{energy function} $E$ for each of the states:
\begin{equation}
E_{\mathbf{W}}\left(t\right)=-\frac{1}{2}\sum_{i=1}^{N}\sum_{j=1}^{N}w_{ij}s_{i}\left(t\right)s_{j}\left(t\right)-\sum_{i=1}^{N}s_{i}\left(t\right)I_{i}.\label{eq:E_W}
\end{equation}
When the connection strengths of $\mathbf{W}$ are symmetric (i.e., $w_{ij}=w_{ji}$)\footnote{\citet{xu_asymmetric_1996} showed that Eq.~(\ref{eq:E_W}) can be modified to accommodate asymmetric weights $\mathbf{W}$, which \citet{xu_asymmetric_1996} argue are both more natural for physiological reasons, and under certain conditions can achieve better performance than the symmetric case.}, and the diagonal elements equal to zero (i.e., $w_{ii}=0$), it can be shown that Eq.~(\ref{eq:state_update}) guarantees that the dynamics of the system will proceed such that the system will reach a stable state at a local minimum (also known as \Hl[r2]{an} \textit{attractor}) of the energy function $E$ in Eq.~(\ref{eq:E_W}) \citep[p.347]{rojas_neural_1996}. A common metaphor used to describe this, is to represent the $N-$dimensional state space (also known as \textit{configuration space}) of the system
as an energy landscape \citep{jones_evolutionary_1995} defined by weights $\mathbf{W}$ and Eq.~(\ref{eq:E_W}). The dynamics of the system are then described by the movement of a dropped ball, that is guided by a landscape of peaks and ridges, that eventually lands in some valley (Fig.~\ref{fig:classical-hns-dyn}), where the valleys are the attractors of the system\footnote{\citet{alphaphoenix_this_2024} offers a comprehensive explanation of thinking in higher dimensions, detailing the concept of higher-dimensional configuration. For a detailed technical explanation of the convergence of the \gls{hn} see \citet{bruck_convergence_1990}.}. This spontaneous convergence to a dynamical attractor is one of the main aspects of \glspl{hn} that allows them to operate in two modes for different purposes, as \emph{associative memory} for pattern recall (\citeauthor{hopfield_neural_1982}, \citeyear{hopfield_neural_1982}; Sec.~\ref{subsec:Pattern-recall}) or as a solver for constraint \emph{optimization} problems (\citeauthor{hopfield_neural_1985}, \citeyear{hopfield_neural_1985}; Sec.~\ref{subsec:Optimization}). The key difference between these two typical \Hl[r1]{modes} of \glspl{hn} lies in how we define the initial weights $\mathbf{W}$. \Hl[r1]{In the \emph{associative memory} mode, we encode in the weights $\mathbf{W}$ the explicit information of the attractor space (i.e., the memories of the network), whereas in the  \emph{constraint optimization} mode, we only have implicit information of the attractor space (i.e., the constraints), and the network is used to compose the sought-after information (i.e., the solution to the problem). Using the energy landscape metaphor, in the first case, we know the locations of the valleys and construct the landscape around them by explicitly encoding their positions. In the second case, we only know certain properties of the valleys (i.e., the constraints), but their exact locations remain unknown. In both cases, the energy landscape is static, meaning that reaching a desired valley (attractor) depends heavily on the initial position of the ball (the network's starting state).}

\Hl[r1]{As was mentioned in Sec.~\ref{sec:Background}, in} this paper, we focus on a third operational mode, \emph{self-optimization} (\citeauthor{watson_optimization_2011}, \citeyear{watson_optimization_2011}\Hl[r1]{, Sec.~\ref{sec:The-Self-Optimization-model}). This mode combines elements of both: we encode in the weights $\mathbf{W}$ the implicit information of the attractor space (i.e., the constraints), and the network generalizes over the attractor space, yielding a global minimum. Thus, contrary to the previous two modes, the energy landscape of the \gls{so} model is constantly changing (Fig.~\ref{fig:so-dyn}).} In the following two sections, we will describe the two operational modes - associative memory and optimization - as basis for introducing the third mode in Sec.~\ref{sec:The-Self-Optimization-model}. \Hl[r1]{The novel contribution of this work is to argue for and present the capacity of the \gls{so} model to exhibit creativity (Sec.~\ref{sec:CC} and Sec.~\ref{sec:SO-and-CC}, respectively).}
\glsreset{hn}
\glsreset{so}

\subsection{HNs as Associative Memory\label{subsec:Pattern-recall}}

In the seminal \citeyearpar{hopfield_neural_1982} paper, \citeauthor{hopfield_neural_1982} showed that patterns can be stored as memories in the weights $\mathbf{W}$ of the \gls{hn} to form an attractor in the dynamical system. \cite{hopfield_neural_1982} showed that we can memorize patterns in $\mathbf{W}$ using a \textit{Hebbian learning} rule\footnote{The rule is named so because of the similarity between Eq.~(\ref{eq:W_hebb}) and the postulate made by \citet[p.~28]{hebb_organization_1949} that synapses increase in strength only if the post-synaptic neuron is fired by some other input at about the same time as a pre-synaptic impulse occurs. This postulate is often summarized as ``neurons that fire together, wire together'' \citep{kuriscak_biological_2015}.}
\begin{equation}
w_{ij}=\begin{cases}
\sum_{k=1}^{M}z^{i}_{k}z^{j}_{k} & i\neq j\\
0 & i=j
\end{cases},\label{eq:W_hebb}
\end{equation}
for $i,j=1,...,N,$ where \Hl[r1]{$N$ is the number of nodes},  \Hl[r2]{$z^{i}_{k}$ and $z^{j}_{k}$} denote the $i$-th and $j$-th component \Hl[r2]{of the $k$-th pattern} of the vector \Hl[r2]{$\mathbf{z}_{k}$} (the pattern we want to store), and $M$ is the total number of patterns. For instance, if we would like to store the patterns of the letters `A' and `B' in a \gls{hn}, we would represent these patterns with the binary vectors \Hl[r2]{$\mathbf{z}_{A}$ and $\mathbf{z}_{B}$}, and use Eq.~(\ref{eq:W_hebb}) to create the weight matrix $\mathbf{W}$. Patterns `A' and `B' then would form the attractors of the \gls{hn} (valleys in Fig.~\ref{fig:classical-hns-dyn}). We can then present a partial \Hl[r1]{(e.g., noisy or corrupt)} pattern $\mathbf{S}_{i}$ (represented by the ball in Fig.~\ref{fig:classical-hns-dyn}) to the network, and the dynamics set by Eq.~(\ref{eq:state_update}) will update the nodes of the network to correct the errors such that eventually the system will converge to the state of the stored memory that the partial pattern represents. The number of memories that could be stored in a \gls{hn} with no errors in recall was shown initially \citep{hopfield_neural_1982,amit_storing_1985} to be limited to $M^{\mathrm{max}}\simeq0.14N$ memories.
\Hl[r2]{Hebbian learning may result in a formation of \glspl{sm}, i.e., additional stable attractors that are not part of the set of desired memories, and storing memories above this threshold may cause these \glspl{sm} to interfere with pattern recall}
\citep{hopfield_unlearning_1983,amit_spin-glass_1985,montgomery_evaluation_1986,rojas_neural_1996}. \citet{krotov_dense_2016} showed that this storage limitation can be alleviated if the standard energy function Eq.~(\ref{eq:E_W}) is modified to have higher (than quadratic) order interactions between the nodes. The formation of \glspl{sm} however can have a paradoxical benefit in the context of optimization, as is discussed in Sec.~\ref{sec:The-Self-Optimization-model}.

\subsection{HNs for Optimization \label{subsec:Optimization}}

In another seminal paper, \citet{hopfield_neural_1985} showed that, if the connections of the \gls{hn} correspond to the constraints of an optimization problem, then the natural dynamics of the system is to converge to a stable state that will correspond to a locally optimal solution to that problem. They illustrated this on the \gls{tsp}: given a list of cities and their pairwise distances, the task is to determine the shortest possible route for the salesperson to visit each city exactly once. 
\Hl[r1]{To make the \gls{hn} converge to a solution (i.e., the shortest path) \citeauthor{hopfield_neural_1985} formulated the problem in terms of desired optima.} For $n$ cities, $N=n^{2}$ nodes were chosen to represent a route, such that every group of $n$ nodes represents the position of a particular city in the route. For example, if for some group of four cities $A,B,C$ and $D$, the shortest possible path is $B\rightarrow D\rightarrow A\rightarrow C$, the desired final state of the network would be represented as a vector of states $\mathbf{v}=\left(0010100000010100\right)$, \Hl[r1]{or in its permutation matrix form}\footnote{\Hl[r1]{This permutation matrix form should not be confused with a graph's adjacency matrix. Traditional methods (e.g., branch-and-bound, dynamic programming) formulate the problem of the \gls{tsp} as a graph traversal problem, where the pairwise distances between the cities are explicitly encoded in the graph’s adjacency matrix \citep{applegate_traveling_2006}. However, a \gls{hn} is an \gls{rnn} that evolves toward a stable state that minimizes an energy function. Since an adjacency matrix representation alone does not inherently enforce the one-city-per-position and the one-visit-per-city constraints, \citeauthor{hopfield_neural_1985} decided to use an encoding that directly represented the problem constraints in the network's weights so that the network dynamics would naturally converge to a valid \gls{tsp} route.}}
\begin{equation}
\begin{array}{cc}
 & \begin{array}{ccccc}
 &  &  & \leftarrow i\rightarrow\end{array}\\
\begin{array}{c}
\\
\uparrow\\
x\\
\downarrow\\
\\
\end{array} & \begin{array}{cc}
 & \begin{array}{cccc}
1 & 2 & 3 & 4\end{array}\\
\begin{array}{c}
A\\
B\\
C\\
D
\end{array} & \left(\begin{array}{cccc}
0 & 0 & 1 & 0\\
1 & 0 & 0 & 0\\
0 & 0 & 0 & 1\\
0 & 1 & 0 & 0
\end{array}\right)
\end{array}
\end{array},\label{eq:TSP_4cities}
\end{equation}
where the row subscript $x$ stands for the city name, and the column subscript $i$ for the position in the route, such that $v_{xi}=1$ (i.e., the neuron is ``on'') would mean that city $x$ is visited in position $i$ of the route. 
Crucially for our later investigation of creativity, \Hl[r1]{the formulation of the problem} does not denote the actual solutions, but the requirements that a solution must fulfill (e.g., optimizing a certain function within constraints).
Using this representation, the \gls{tsp} requirements can be expressed as two constraints to ensure that the output vector $\mathbf{v}$ composed of elements $v_{xi}$ corresponds to a valid tour:
\begin{enumerate}
\item There must be one and only one neuron that is ``on'' in each row of $v_{xi}$. Informally, the salesperson must visit each city, but only once.
\item There must be one and only one neuron that is ``on'' in each column of $v_{xi}$. Informally, the salesperson can't be at two places at once.
\end{enumerate}
Using these constraints, the energy function (\ref{eq:E_W}) can be reformulated such that it will be minimized for a state that satisfies these constraints \Hl[r2]{and is the shortest route}:
\begin{align}
E= & \frac{A}{2}\sum_{x=1}^{N}\sum_{i=1}^{N}\sum_{j\neq i}^{N}v_{xi}v_{xj}+\frac{B}{2}\sum_{i=1}^{N}\sum_{x=1}^{N}\sum_{y\neq x}^{N}v_{xi}v_{yi}\nonumber \\
 & +\frac{C}{2}\left(\sum_{x}\sum_{i}v_{xi}-n\right)^{2}\nonumber \\
 & +\frac{D}{2}\sum_{x}\sum_{y\neq x}\sum_{i}d_{xy}v_{xi}\left(v_{y,i+1}+v_{y,i-1}\right),\label{eq:E_TSP}
\end{align}
where $A$, $B$, and $C$ are positive constants, and $d_{xy}$ are the pairwise distances between the cities. The first two terms are zero iff there is no more than one neuron ``on'' in a row or column, respectively, and the third term is zero iff there are $n$ entries in the entire matrix. Together, these enforce the two constraints above. The fourth term is added to favor short paths. The corresponding weight matrix $\mathbf{W}$ can then be extracted through a term-by-term comparison with Eq.~(\ref{eq:E_W}). Different than pattern recall (Sec.~\ref{subsec:Optimization}), here the system's dynamical attractors represent (partial) solutions to the optimization problem, with potentially several global attractors representing optimal solutions with equally short paths. 
\glsreset{so}
The mathematical procedure of translating the constraints of the \gls{tsp} to the variables of a \gls{hn} is beyond the scope of this paper and covered in detail by \citet{aiyer_theoretical_1990}. Here we want to point out that the \gls{tsp} problem is considered NP-complete\footnote{The ``NP-complete'' stands for ``nondeterministic polynomial time complete'', and it means in simplified terms that the problem is verifiable (but not necessarily solvable) in polynomial time.} \citep[p.434]{hopcroft_introduction_2007}, which means it is computationally intractable for large sizes of $n$. Similar to pattern recall, what makes an optimization problem such as the \gls{tsp} so hard to solve is the presence of vastly many more local minima than the global attractors we want the system to find \citep[p.369]{rojas_neural_1996}. \Hl[r2]{As \citet{hopfield_neural_1985} report, for a 10-city problem their network produced one of the 2 shortest paths only about 50\% of the trials.} To deal with this, a myriad of \Hl[r2]{approximation} methods were developed since \Hl[r2]{in an attempt to find solutions to the \gls{tsp} and other} optimization problems
with derivatives of \glspl{hn} or other types of networks, using methods like Genetic algorithms or heuristic search \citep{rojas_neural_1996}.
The advantage of the \gls{so} model, which is the subject of next section, is that it retains both the elegant mathematical framework of \glspl{hn} and their biological plausibility. Unlike classical \glspl{hn}, however, the \gls{so} model can manipulate the local minima within the ``bumpy'' high dimensional surface  of \glspl{hn} to find more optimal solutions \citep{watson_optimization_2011}.

\section{The Self-Optimization Model: Leveraging the Power of Associative Memory for Optimization\label{sec:The-Self-Optimization-model}}
\glsreset{hn}
\glsreset{so}
\glsreset{sm}
As mentioned in Sec.~\ref{subsec:Pattern-recall}, the formation of \glspl{sm} in a \gls{hn} with Hebbian learning is usually considered undesirable, since it degrades the memory performance of the network \citep{montgomery_evaluation_1986}. Several researchers in the 90s, however, pointed out that Hebbian learning opens up additional avenues for investigating \glspl{hn} beyond their memory capabilities.
\citet{fontanari_generalization_1990} pointed out that above a certain critical number of patterns, the network enters a regime of \textit{generalization}, where it can capture the underlying statistical structure of the patterns that the system is trained on. \citet[p.~I-25]{jang_conceptual_1992} further emphasized that this formation of meaningful \glspl{sm} bears parallels to the creation of conceptual knowledge by the brain, ``loosely speaking \emph{creative thinking}'', which touches on the topic of this article. Building on this work and the dependence of ``the deeper minimum \ensuremath{\leftrightarrow} the larger the basin of attraction \ensuremath{\leftrightarrow} the larger the probability to get to this minimum\textquotedblright{} \citep[p.~97]{kryzhanovsky_binary_2008}, Watson and colleagues came up with a crucial insight: one can harness the generalization effect of Hebbian learning in \glspl{hn} to significantly boost the optimization performance of \glspl{hn} \citep{watson_effect_2009,watson_optimization_2011,watson_transformations_2011}.
In this section, we will describe the basics of \Hl[r1]{their model, which we refer to as} the \gls{so} model, as well as how it has been used so far.

\subsection{The basics of the Self-Optimization Model}

At the basis of the \gls{so} model is a \gls{hn} (Sec.~\ref{subsec:HN-dynamics}), with the state of the system defined by a vector $\mathbf{S}\left(t\right)=\left\{ s_{1}\left(t\right),...,s_{N}\left(t\right)\right\}$ in a bipolar representation that is updated by Eq.~(\ref{eq:state_update}). The initial weights $\mathbf{W}_{0}$ represent a constraint optimization problem (Sec.~\ref{subsec:SO-weights}). Here and throughout, we assume symmetric weights, though the model also applies to asymmetric cases \citep{zarco_self-optimization_2018}. The system is initialized to a random state and allowed to converge to a local attractor. Similarly to the typical optimization case discussed in Sec.~\ref{subsec:Optimization}, at each time step, a node is randomly chosen and updated by Eq.~(\ref{eq:state_update}). Differently though, the weights are then updated by Hebbian learning (i.e., the learned weights $\mathbf{W}_{\mathrm{L}}$), following the rule
\begin{equation}
w_{ij}^{\mathrm{L}}\left(t+1\right)=w_{ij}^{\mathrm{L}}\left(t\right)+\alpha s_{i}\left(t\right)s_{j}\left(t\right),\label{eq:w_update}
\end{equation}
where $w_{ij}^{\mathrm{L}}\left(t=0\right)=w_{ij}^{\mathrm{0}}$, and $\alpha>0$ is a learning rate constant. After performing $T$ steps the state is reset again to a new random state. The weight matrix $\mathbf{W}_{\mathrm{L}}$ is not reset. In total, the procedure is repeated for $R$ resets. To assess how the system performs with regards to the initial problem, the energy of the system is always computed with regards to the initial weight matrix $\mathbf{W}_{0}$, but the dynamics and the state update depend on the learned weight matrix $\mathbf{W}_{\mathrm{L}}$. Very importantly, since the energy landscape of the system is defined by the learned weights $\mathbf{W}_{\mathrm{L}}$ through Eq.~(\ref{eq:E_W}), every modification of the weights results in the modification of the entire landscape (Fig.~\ref{fig:so-dyn}) and, as we will see in Sec.~\ref{sec:SO-and-CC}, the learning rate $\alpha$ has a crucial role on the resulting landscape.

\subsection{Initial Weights and the Range of Applications\label{subsec:SO-weights}}

\begin{figure}[t!]
     \centering
     \begin{subfigure}[b]{0.49\textwidth}
         \centering
         \includegraphics[width=\textwidth]{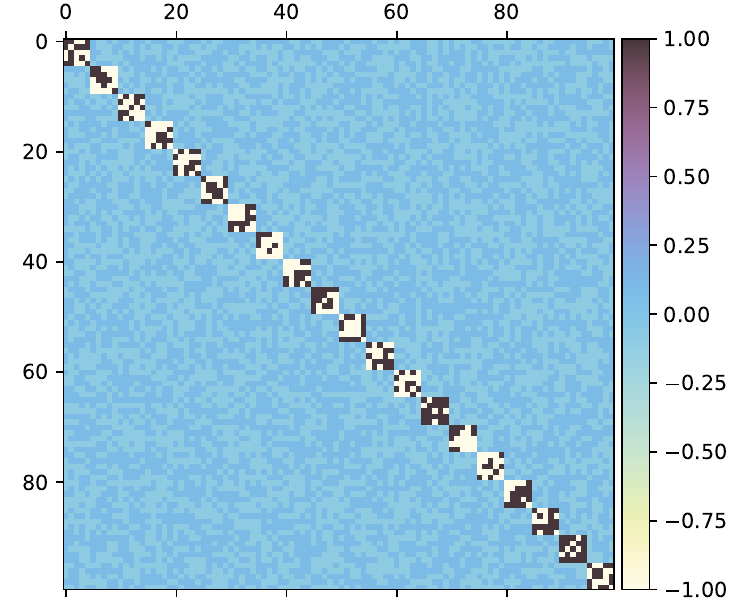}
         \caption{Initial weights}
         \label{fig:W-init}
     \end{subfigure}
     \hfill
     \begin{subfigure}[b]{0.49\textwidth}
         \centering
         \includegraphics[width=\textwidth]{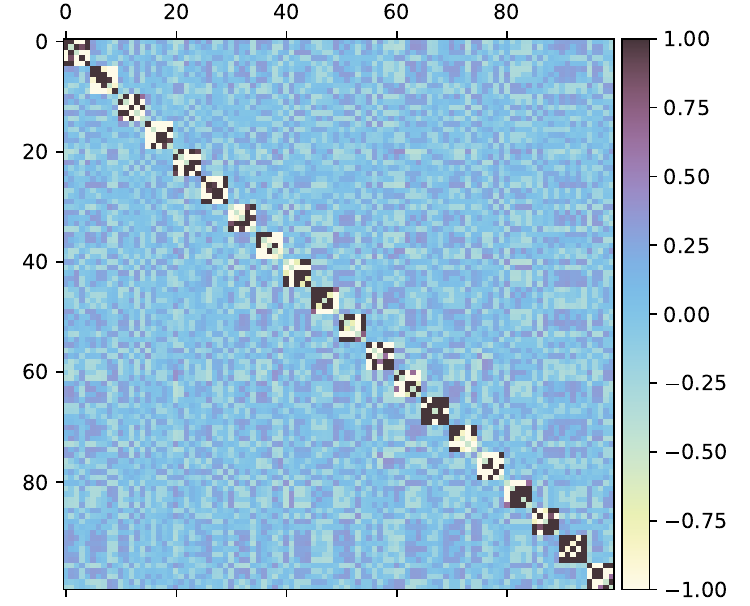}
         \caption{Final weights}
         \label{fig:W-fin}
     \end{subfigure}
        \caption{\Hl[r1]{An illustration of the \gls{so} weight matrix and the effect of learning. It shows a} symmetric modular connectivity weight matrix, Eq.~(\ref{eq:modular_W}), (a) before and (b) after learning ($\alpha=5\mathrm{e}{-7}$) for a system of size $N=100$. Initial weights have 20 modules of size $k=5$, intra-module weights \Hl[r2]{$p$} set at random to either 1 or -1, and inter-module weights set at random to either 0.1 or -0.1.}
        \label{fig:W}
\end{figure}

\citet{watson_effect_2009} chose specific weights $\mathbf{W}$ for illustrative purposes of the \gls{so} procedure: randomly generated weight matrices with different structures, representing different
scenarios. One such structure is modularity, which is a common characteristic of natural dynamical systems \citep{watson_modular_2005}. Modular connectivity weight matrix (Fig.~\ref{fig:W}) with parameterized strength of inter-module connections is given by
\begin{equation}
w_{ij}=\begin{cases}
\pm 1 & \left\lfloor \frac{i}{k}\right\rfloor =\left\lfloor \frac{j}{k}\right\rfloor \\
\pm p & \mathrm{otherwise}
\end{cases},\label{eq:modular_W}
\end{equation}
where $k$ is the size of the modules, $\left\lfloor x\right\rfloor $ is the floor function of the value $x$, and $p$ is the inter-module weight, and the sign is chosen randomly. Modular weights as in Eq.~(\ref{eq:modular_W}) represent a scenario \Hl[r2]{, in which on top of the} two global minima, \Hl[r2]{there are} $2^{N/k}$ deep local optima. \Hl[r2]{The strong intra-module weights combined with the weak inter-module weights cause the local minima to be far apart with a shallow gradient in between}, such that the Hopfield update (Eq.~\ref{eq:state_update}) alone cannot escape the local \Hl[r2]{optima}. Thus finding the global optimum through choosing random initial states and Hopfield updates alone is exceedingly unlikely \citep{watson_effect_2009}. This structure of weights was used to abstract and simulate various complex adaptive systems \citep{watson_optimization_2011}, such as sociopolitical networks \citep{froese_can_2014,froese_modeling_2018} or social coordination system of driving conventions across countries \citep{tissot_ability_2024}. 

\citet{watson_optimization_2011} pointed out that the weights~(\ref{eq:modular_W}) can also represent a weighted-Max-2-SAT problem, which prompted \citet{weber_use_2023-1} to show how one could, in principle, \Hl[r2]{convert any \gls{sat}}\footnote{SAT \Hl[r2]{(satisfiability problem) refers to a class of combinatorial problems in propositional logic. MaxSAT (maximum satisfiability) is an optimization problem that aims to maximize the number of satisfied clauses in a Boolean formula where no solution exists that satisfies all clauses simultaneously.} Max-k-SAT \Hl[r2]{is a MaxSAT variant} with a maximum of $k$ literals per clause. \Hl[r2]{MaxSAT is a special case of Weighted-Max-SAT, where all clause weights are equal to 1 \citep{biere_handbook_2009}.}} \Hl[r2]{problem into the weights of a \gls{hn} and apply the \gls{so} model in an attempt to find a solution.}
\citeauthor{weber_use_2023-1} demonstrate that on two classical examples: the Liars, and the map coloring problem. \Hl[r2]{While this conversion is possible, it does not imply that the \gls{so} model can efficiently solve the \gls{sat} problem in the sense of finding the global minimum in polynomial time - there is no guarantee of convergence to the optimal solution.}\footnote{\Hl[r2]{The \gls{so} model can be regarded as an ``incomplete'' solver - it may not find the optimal solution and it does not provide a proof of unsatisfiability for unsatisfiable instances, unlike complete, state-of-the-art \gls{sat} solvers.}} \Hl[r2]{ However, t}he \gls{so} model's ability \Hl[r2]{to potentially find solutions}  \Hl[r2]{to \gls{sat}} problems 
is significant\Hl[r2]{, as} many real-world problems across various scientific fields can naturally be expressed as Max-k-SAT \citep{biere_handbook_2009}. \Hl[r1]{For instance, \gls{sat} methods are used in software verification. Computer software errors may lead to anything from minor issues, to financial losses to even loss of life depending on where the computers are embedded (private home vs communication vs transportation systems). \gls{sat} are useful for software verification because software behavior can be modeled closely to how it actually runs on hardware by modeling bit-vectors as Boolean variables and operations as a set of Boolean functions \citep{biere_handbook_2009}.} 
\Hl[r2]{While the \gls{so} model is not comparable to state-of-the-art \gls{sat} solvers in terms of speed or guaranteed optimality, its significance lies elsewhere: it provides a biologically realistic mechanism with minimal assumptions that can, in some cases, yield solutions to real-world problems. This makes it an important aspect for an \gls{alife} model of minimal creativity.} In addition, focusing on a concrete problem enables us to anchor value as characteristic of creative outcomes, and explore how such outcomes are affected by learning, a topic we will revisit in Sec.~\ref{subsec:novelty-SO}.

Finally, as we have seen, with the classical optimization mode of \glspl{hn} (Sec.~\ref{subsec:Optimization}), one has to adjust the mathematical formula of the energy function (\ref{eq:E_W}) in order to encode the requirements based on the desired end-state. In the \gls{so}, however, this is not necessarily needed. If the original problem can be converted to a list of propositional clauses, then one can use a given straightforward procedure of translating the list to the clauses to the weights and then \Hl[r2]{apply the \gls{so} to} the problem with the regular dynamical equations of \glspl{hn} as described in detail in \citep{weber_use_2023-1}.

\subsection{The Self-Optimization Dynamics\label{subsec:SO-dynamics}}
\begin{figure}[t!]
     \centering
     \begin{subfigure}[b]{0.49\textwidth}
         \centering
         \includegraphics[width=\textwidth]{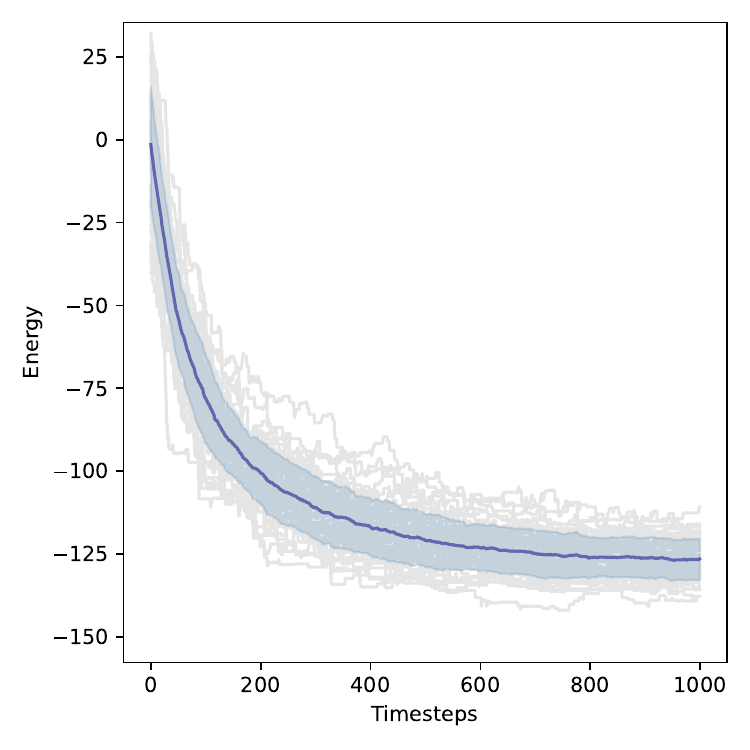}
         \caption{Without learning}
         \label{fig:E_NL}
     \end{subfigure}
     \hfill
     \begin{subfigure}[b]{0.49\textwidth}
         \centering
         \includegraphics[width=\textwidth]{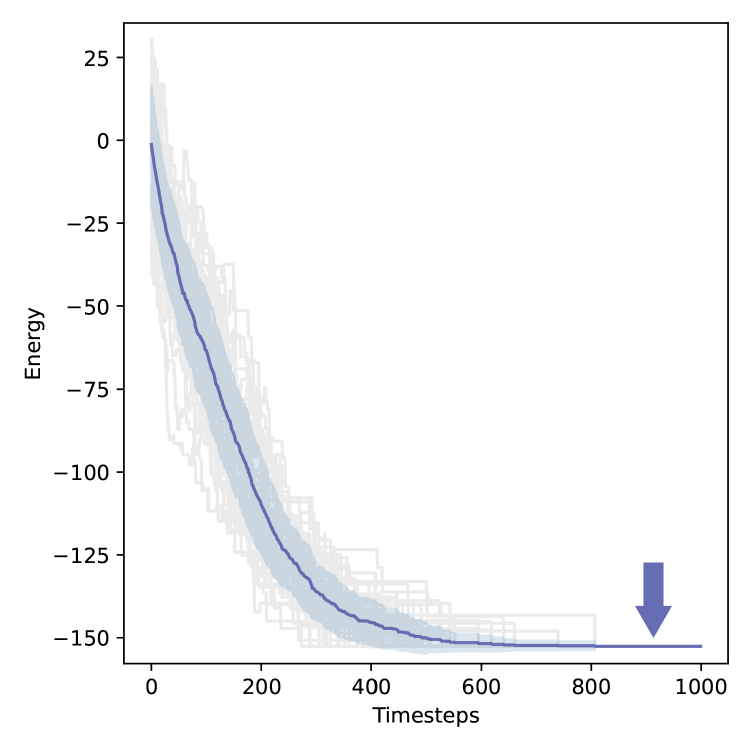}
         \caption{With learning}
         \label{fig:E_L}
     \end{subfigure}
        \caption{Dynamics of the \gls{so} model with and without learning. (a) Without learning. This is equivalent to the dynamics of a regular Hopfield network, where the system will converge to various minima according to the initial state. (b) With learning, Eq.~(\ref{eq:w_update}). The arrow indicates converges to a single lower energy attractor for all  initial random states. On both plots, the energy is computed using Eq.~(\ref{eq:E_W}) for the chosen initial weight matrix $\mathbf{W}_{\mathrm{0}}$ in Fig.~\ref{fig:W-init}, and the state at each step is updated asynchronously according to Eq.~(\ref{eq:state_update}). The plots show the energy for 50 different initial random states each.}
        \label{fig:E_conv}
\end{figure}
Here and throughout all \gls{so} simulations were done with initial weights $\mathbf{W}_{0}$ as in Fig.~\ref{fig:W-init}. Given an optimally chosen learning rate $\alpha$ (more on this in Sec.~\ref{sec:SO-and-CC}) in Eq.~(\ref{eq:w_update}), the system undergoes self-organization and converges to a single lower energy attractor (as shown by the arrow in Fig.~\ref{fig:E_L}), regardless of the initial state of the system and even if the system has not previously visited that attractor before (Fig.~\ref{fig:N_A}). Through repeated energy minimization and gradual adjustments to the connections, the network evolves into an associative memory of its own attractor states. As the weights accumulate, the sizes of the attractor basins also change - some basins expand at the expense of others. This process creates a positive feedback loop that significantly enhances the system's ability to resolve the constraints of the original weight matrix and find the configuration of lower energy states  \citep{watson_optimization_2011}, as schematically depicted in Fig.~\ref{fig:so-dyn}.

Arguably the most important observation here is that, by virtue of Hebbian learning, one can 
\Hl[r2]{exploit the readily available local minima within the problem's state space to widen the attractor basin of the initially hard-to-find global optimum, thereby increasing the likelihood of discovering optimal solutions.}
\Hl[r1]{Given this generalization capacity of the \gls{so} model, } it is tantalizing to ask what it can teach us about (creative) problem-solving in (artificial) life.

\subsection{Minimal Agency in the SO Model}
\label{subsec:so_agency}
We consider the \gls{so} model so useful, as it not only allows us to study creativity (Sec. \ref{sec:SO-and-CC}) but also its relationship with agency as characteristic of life. Both \citet{froese_autopoiesis_2023} and \citet{watson_agency_2024} converged on the insight that the \gls{so} model provides insights into the basis of minimal forms of agency. Briefly, at the core of living systems lies an inherent primordial tension of self-individuation: an organism's far-from-equilibrium embodiment requires it to flexibly alternate between states of relative openness and closedness to material exchanges with the environment over time. This alternation can be interpreted as the ``resets'' in the \gls{so} procedure which, together with accumulation of structural changes in the weight matrix due to Hebbian learning, give rise to a system-level competency to forgo short-term gains (by avoiding local minima) in favor of long-term gains (by getting closer to lower optima). \citet{froese_autopoiesis_2023} utilize this to explore on an abstract level the minimal conditions of adaptive regulation in a living system.

\citet[p.~23]{watson_agency_2024} proposes a ``scale-relative notion of agency'' in a biological system, where a system is considered agential if its constituent parts ``are able to acquire organized relationships among themselves that exhibit goal-directed behavior greater than that which they exhibit as individuals'' (p.~30). At the core is the idea that a fundamental feature of all living organisms is their problem-solving ability in a space of varying possibilities. A goal-directed behavior in the \gls{so} model can be interpreted as improvement in its problem-solving ability (i.e., finding better solutions to combinatorial constraint optimization problems). \citet[p.~30]{watson_agency_2024} describe\Hl[r1]{s} two ways in \Hl[r2]{which} the system in the \gls{so} model learns to be an agent by resolving its own constraints. It can be \textit{conservative} or \textit{innovative}. When the system is conservative, it ``holds on to good states already identified''. When innovative, it promotes ``good states that are novel — that have not previously been visited in past experience''\footnote{This bears close resemblance to the exploration-exploitation trade-off commonly articulated in (computational) reinforcement learning.}. \citet{watson_agency_2024} refers to the latter innovation as a type of \textit{generalization}, i.e., ``the ability to respond correctly to novel inputs or generate (nonrandom) novel patterns''. This last remark suggests that the \gls{so} may also serve as a model of a creative process.

The last observation will become clearer with the introduction of various definitions of creativity in the following section. As our main contribution in this article, we argue that the \gls{so} model constitutes a framework for investigating minimal creativity, and study the dependency of creative outcomes on its learning parameters. We return to the open question of the relationship between life, agency, and creativity later in our discussion (Sec.~\ref{sec:Discussion}).

\section{Creativity: Perspectives and Definitions\label{sec:CC}}

If we want to study creativity in the \gls{so} model, we must first clarify what we mean by ``creativity''. Due to an amalgamation of different concepts in the same word \citep{still_history_2016, green_process_2023}, creativity can be attributed, amongst others, to both the \emph{process} in which someone or something engages in, and the \emph{product} of that process \citep{rhodes_analysis_1961}. \Hl[r1]{The product- and process-based perspectives can be used separately or in conjunction, depending on the context. The former focuses on assessing whether a product can be considered creative and why it might be more creative than another. The latter allows us to determine when a (mental) process can be considered creative. Crucially, sometimes, a creative product may result from a non-creative process, and a creative process might not result in a creative product or no product at all. Thus focusing exclusively on one or the other may limit the broader understanding of creativity. Next,} we discuss these two perspectives on creativity as prerequisite for investigating how the \gls{so} model fits into each in Sec.~\ref{subsec:creativity-product} and \ref{subsec:creativity-process}, respectively. As one of our novel contributions in this article, we also argue in Sec.~\ref{subsec:creativity-process} that the \gls{so} model meets all requirements of \Hl[r1]{the process} definition \Hl[r1]{of creativity }and thus instantiates a creative process.

\subsection{Creativity as a Product\label{subsec:creativity-product}}

Definitions of creativity as \Hl[r2]{an} attribute of a \textbf{product} (``creativeness''; \citeauthor{green_process_2023}, \citeyear{green_process_2023}) are aplenty\footnote{$\,$See \citet{kampylis_redefining_2010} for a review of 42 explicit definitions of creativity from various experts in the field from 1950 to 2009.}. \citet{runco_standard_2012} reviewed many such definitions to identify common core components, and feature them in a now very popular ``standard definition'' of creativity: ``\emph{Creativity requires both originality and effectiveness}''. Despite the proclaimed consensual agreement, however, the definition that citing researchers use often differs substantially. This is likely because the standard definition is ambiguous, since both of its core components are underdefined: ``originality'' may also be labeled as novelty (which can be assessed on a personal or historical level; see \citeauthor{boden_creativity_2015}, \citeyear{boden_creativity_2015}) or a type of variation, innovation, or transformation \citep{stepney_modelling_2021}, and ``effectiveness'' may take the form of value, utility, fitness, or aesthetic pleasure, just to name a few. Following extensive debate supported by \citet{beghetto_dynamic_2019}, \citet{runco_creativity_2019} acknowledges himself that the standard definition refers to the \emph{products} generated at different stages of the creative process. We can therefore conclude, that there is general agreement on what constitutes a \emph{creative outcome}, namely that it is both \emph{novel} and \emph{effective}\Hl[r1]{, but it is not fully self-contained as the concrete meaning of what is novel or effective is context dependent. }Here we will use the criterion of task \emph{appropriateness} instead of effectiveness to emphasize suitability of the product (tangible or intangible) for its original  \Hl[r1]{goal}. Thus the following definition will be used throughout the text:
\begin{definition}[\emph{Creative product}] A product (tangible or intangible) that is both novel and appropriate.\label{Def:Def_product}
\end{definition}
In Sec.~\ref{sec:SO-and-CC} we \Hl[r1]{expand on what `novelty' and `appropriateness' in the context of products of the \gls{so} model mean, and} show that the \gls{so} model can generate different outcomes, either novel or appropriate or both, and discuss the conditions under which creative outcomes emerge. Definition (\ref{Def:Def_product}) only captures the product as one perspective on the creativity of the \gls{so} model. Next, we introduce a second definition on the process.

\subsection{Creativity as a Process\label{subsec:creativity-process}}

\citet{corazza_potential_2016} notes that the standard definition, and other similar definitions \citep{kampylis_redefining_2010}, that are solely based on fixed criteria of \emph{static creative achievement}, cannot capture the dynamic process that may or may not lead to a creation and the sequence of \emph{inconclusive outcomes} that precede the final achievement. Creativity as a dynamic phenomenon was further investigated by \citet{beghetto_dynamic_2019}. And more recently, \citet{green_process_2023} provided for the first time a definition of the creative \textbf{process}. They argue that, although many models of the creative processes exist, there is no one definition that ``distill{[}s{]} the minimal set of criteria that constitute necessity and sufficiency'' to be general enough to cover creativity across many forms \citep[p.~2]{green_process_2023}. They formulate the process definition of creativity as 
\begin{definition}[\emph{Creative process}] Internal attention constrained by a generative goal.\label{Def:Def_Green2023}
\end{definition}
It encompasses three different criteria, \Hl[r1]{internally-directed attention (Sec.~\ref{subsubsec:attention}), goal constraint (Sec.~\ref{subsubsec:constrained-goal}), and generative goal (Sec.~\ref{subsubsec:generative-goal}),} which \citet[p.~11]{green_process_2023} claim to be necessary and sufficient for a process to be called creative. 

\Hl[r1]{The process definition is useful in that it abstracts from the many, more concrete models of creative processes, providing more general, necessary and sufficient criteria that all such models must implement. As such, it enables us in studying creativity in the \gls{so} model without residing to a specific application context or biological substrate; this general application also enables future work in interpreting the \gls{so} model with respect to more concrete and context-specific models of the creative process.

Although having only been recently introduced, the validity of the process definition is supported by its derivation from a priori theoretical premises and a posteriori alignment with empirical evidence from the neuroscience of creativity. 

While this supports the necessity of the conditions presently comprised within the process definition, we remain critical on whether these are sufficient to describe any instance of creative processes. In particular, some researchers might consider further criteria, such as intentionality, to move away from a potentially trivializing view of creativity as `mere generation' \citep{ventura_mere_2016}. This particularly concerns how the value, e.g., constraints on the goal state are derived from, is grounded. We believe that re-evaluating the \gls{so} model's validity to represent different types of creative processes in response to potential revisions of the process definition an important and interesting avenue of future work.}

We next briefly expand on each criterion, and argue how it can be implemented in the \gls{so} model.

\subsubsection{Internally-directed attention\label{subsubsec:attention}}

The first criterion of \emph{internally-directed attention} implies that creativity necessarily operates over information that has some internal representation (e.g., content retrieved from memory) and not directly over external stimuli (e.g., sensory information such as features of an object or a sound). 
\Hl[r2]{However, while }\citet{green_process_2023} argue that attention to external stimuli may not be necessary for creativity, separating external attention from internal attention (and other types of attention) may not be practically possible. \citet{narhi-martinez_attention_2023} argue that in real life, there is rarely just one single target of attention (external or internal). 
For this reason, \citet{narhi-martinez_attention_2023} propose to define attention as \emph{a multi-level system of weights and balances}.\footnote{\citet{krauzlis_attention_2014} posit a somewhat similar notion by proposing that attention arises as a byproduct of decision making that depends on the current state of the animal and its environment. Each candidate state is determined from a competition between the weights applied to the sensory and non-sensory inputs.} 


\Hl[r1]{We apply \citeauthor{narhi-martinez_attention_2023}'s \citeyearpar{narhi-martinez_attention_2023} hierarchical definition to the \gls{so} model. Following this interpretation, the original weights, $\mathbf{W}_{\mathrm{0}}$, defined by the external problem, may represent external attention. Internal attention in the model, on the other hand, can be characterized by the learned weights, $\mathbf{W}_{\mathrm{L}}$. As soon as the system starts to learn, it starts to prioritize (`attend') to some things more than others, thereby creating its own internal representation of the problem (resulting in the modification of the energy landscape, Fig.~\ref{fig:classical-hns-dyn}). In other words, by learning, the agent transitions from relying solely on an externally defined problem to developing internal preferences on the basis of accumulated memory.}

On a higher level of interpretation, \citet{narhi-martinez_attention_2023}'s definition of attention aligns with the \gls{so} model in two ways. First, the temporal aspect of attention (i.e., bias towards past experience or future goals) can be represented as a ``delayed gratification'' mentioned in Sec.~\ref{subsec:so_agency}, where \Hl[r2]{by attending to its internal preferences} the system follows dynamical trajectories that forego short-term gains in favor of long-term gains \citep{watson_agency_2024}. Second, the multi-level system aspect aligns with \Hl[r1]{\citeauthor{watson_agency_2024}'s} notion of agency of a biological system, that exists on multiple scale\Hl[r2]{s}, at a lower scale of its parts, and a higher scale of agency of the system as a whole.

\subsubsection{Goal constraint\label{subsubsec:constrained-goal}}


\Hl[r1]{The second criterion is that throughout the process the attention is constrained by goal state parameters. The \textit{goal state} presumes any end-state as long as it is not fully accessible during the process (see Sec.~\ref{subsubsec:generative-goal})}.
\Hl[r1]{This criterion} reflects the theoretical assumption that creativity is not just a random, but a meaning-based process. Without constraints, ``ideation would be a means without an end'', more representative of intrusive thoughts or disordered associations \citep[p.~12]{green_process_2023}. Thus, fitting the goal state parameters can be considered as a meaningful process (meaningful to the creator in relation to the \Hl[r2]{constraints} of the goal state). In this sense, constraints of a goal is a process-based view of the product-based requirement of appropriateness. 

In the \gls{so} model, the goal state is the state that satisfies \Hl[r2]{most of} the problem's constraints (i.e., the goal) by resolving the maximum tension within the network. The constraints of the goal are embedded explicitly into the weight matrix $\mathbf{W}$ (see Sec.~\ref{subsec:SO-weights}). These ``desired optima'' \citep[p.~141]{hopfield_neural_1985} match the definition of goal constraints as ``any attributes of representations that are preferable to the individual in the process of generating and selecting candidate representations'' \citep[p.~12]{green_process_2023}. These weights are not to be confused with the weights that can be added on top of the constraints to put emphasis on particular \Hl[r1]{external} target of attention as mentioned above (see \citet{weber_use_2023-1} for an example that distinguishes between these two types of weights). 

As mentioned, the implication of the constrained goal is that the process is not random \Hl[r2]{but purposeful}. This matches the \emph{non-randomness} requirement for creative autonomy put forward by \citet{jennings_developing_2010}. \Hl[r2]{The idea is that a computational procedure would likely not be deemed creatively autonomous if it did not yield creative products with a probability significantly above chance (i.e., a random number generator is not creative). To drive the point home, consider a situation where a person creates a thousand random different products with no underlying strategy, and one of them happens to be creative. We typically would not consider this process very creative.} In Sec.~\ref{subsec:above-chance} we demonstrate that the \gls{so} model generates creative products with a probability far exceeding chance.

\subsubsection{Generative goal\label{subsubsec:generative-goal}}

The final criterion requires that the goal be \emph{generative}. A generative goal implies that the goal state is not already held in memory, otherwise reaching it would be just considered as memory retrieval. \cite{green_process_2023} comment that retrieval can be part of the process, but the goal state must \emph{change} with respect to the retrieved information. 

To understand how the goal in the \gls{so} model is generative, let us recall again the notion of desired optima in the regular optimization mode of the \gls{hn}  (Sec.~\ref{subsec:Optimization}) and use the \gls{tsp} as example task. As an experimenter, we do not know ahead of time the optimal traveling route (i.e., the goal state). All we know is the constraints that must be satisfied for us to consider a goal to be achieved, i.e., the shortest route visiting all cities only once. The construction of the energy function in Eq.~(\ref{eq:E_TSP}), here exemplified on the \gls{tsp}, ensures that the goal state is present in the state space of all possible solutions. In practice, however, it is unlikely for the system to find it by chance. As we shall see in Sec.~\ref{sec:SO-and-CC}, the goal state is well outside of the range of states already held in the memory of the system (i.e., it is generative), and the learning in the \gls{so} model dramatically improves the odds of reaching it. As we will see in the next section, a generative goal on its own, however, does not guarantee that the final state (i.e., the product) would be appropriate, which, as shown in Sec.~\ref{sec:SO-and-CC}, substantially depends on the learning rate. There is no issue with this however, since \citet[p.~15]{green_process_2023} note that the process-based definition ``does not require that the product is actually appropriate (or affordant) even to the individual'' for the process to be considered creative.

\section{SO Model as a Model for Studying Creativity\label{sec:SO-and-CC}}

After establishing that the \gls{so} model instantiates a creative process by fulfilling all of Green et al's \citeyearpar{green_process_2023} criteria (Sec.~\ref{subsec:creativity-process}), we next employ the product definition (Sec~\ref{subsec:creativity-product}) to argue that an \gls{so} model's intermediate steps and solutions can qualify as creative products, and the model allows us to study the exact circumstances in which such creativity emerges. 

The final \textit{products} of the \gls{so} model are the attractor states the system ends in by the end of the learning stage. We classify these outcomes according to the criteria for creative products (Sec.~\ref{subsec:creativity-product}), namely \textbf{novelty} and \textbf{appropriateness}\Hl[r2]{, with respect to the outcomes of the dynamics before learning}. The final state is judged as novel if the system had not visited that attractor within the distribution of energies before learning. Appropriateness is harder to define a priori given that we are looking at a randomly generated weight matrix (Fig.~\ref{fig:W-init}; we get back to this point in Sec.~\ref{subsec:novelty-SO}). For now, we define a state to be appropriate if 
\Hl[r2]{a) it converged to an attractor, and b) that attractor is lower in energy than the mean attractor states within the distribution of energies before learning. The former is a requirement that the system indeed selected a final solution. The latter is an attribution of value to that solution. It} assumes that lower energy states have less tension in the system, which means satisfaction of a higher number of constraints, thus being more appropriate for the task at hand.

\begin{figure}[t!p]
     \centering
     \begin{subfigure}[b]{0.49\textwidth}
       \centering
       \makebox[30pt]{}\textbf{Not Novel}
       
     \end{subfigure}
     \hfill
     \begin{subfigure}[b]{0.49\textwidth}
       \centering
       \makebox[30pt]{\textbf{Novel}}
     \end{subfigure}
     \begin{subfigure}[b]{0.49\textwidth}
        \makebox[0pt][r]{\makebox[30pt]{\raisebox{0.5\textwidth}{\rotatebox[origin=c]{90}{\textbf{Not appropriate}}}}}%
         \includegraphics[width=\textwidth]{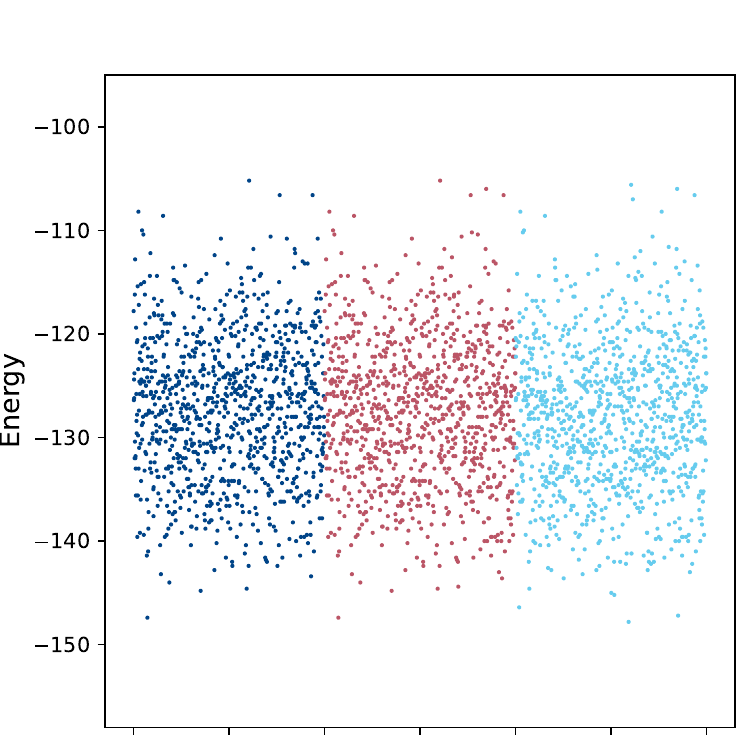}
         \caption{\Hl[r2]{$\alpha=3\mathrm{e}{-8}$}}
         \label{fig:NN_NA}
     \end{subfigure}
     \hfill
     \begin{subfigure}[b]{0.49\textwidth}
         \centering
         \includegraphics[width=\textwidth]{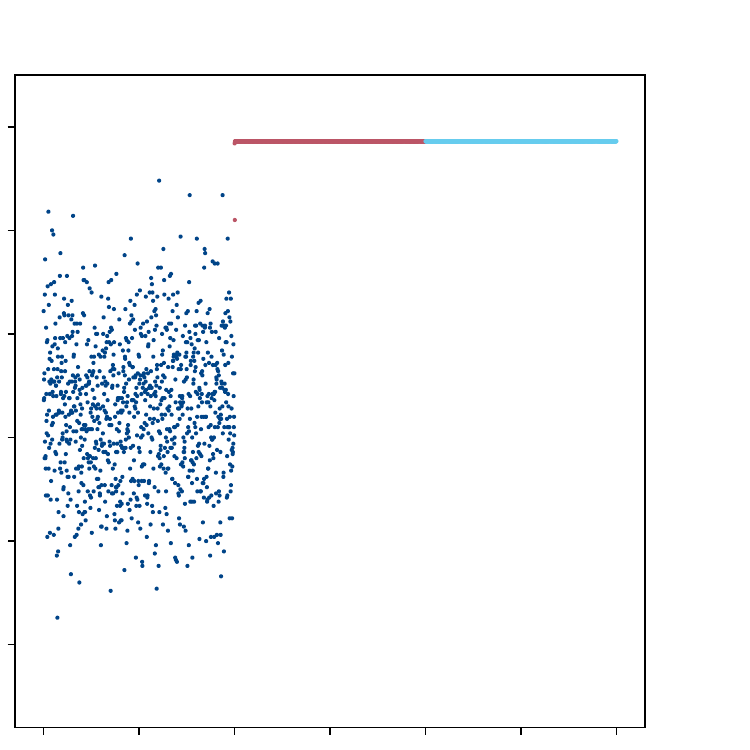}
         \caption{$\alpha=5\mathrm{e}{-5}$}
         \label{fig:N_NA}
     \end{subfigure}
     \begin{subfigure}[b]{0.49\textwidth}
         \makebox[0pt][r]{\makebox[30pt]{\raisebox{0.5\textwidth}{\rotatebox[origin=c]{90}{\textbf{Appropriate}}}}}
         \includegraphics[width=\textwidth]{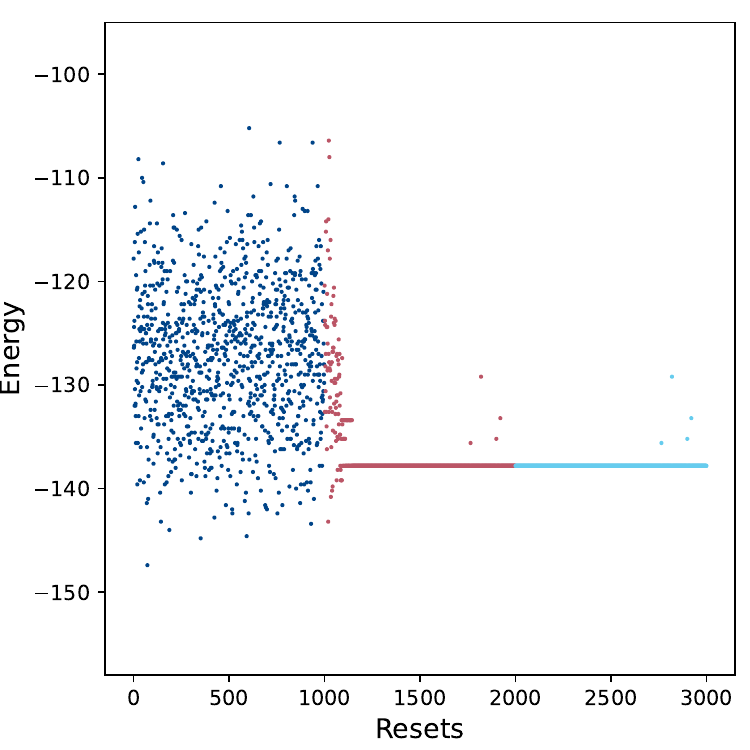}
         \caption{\Hl[r2]{$\alpha=3\mathrm{e}{-6}$}}
         \label{fig:NN_A}
     \end{subfigure}
     \hfill
     \begin{subfigure}[b]{0.49\textwidth}
         \centering
         \includegraphics[width=\textwidth]{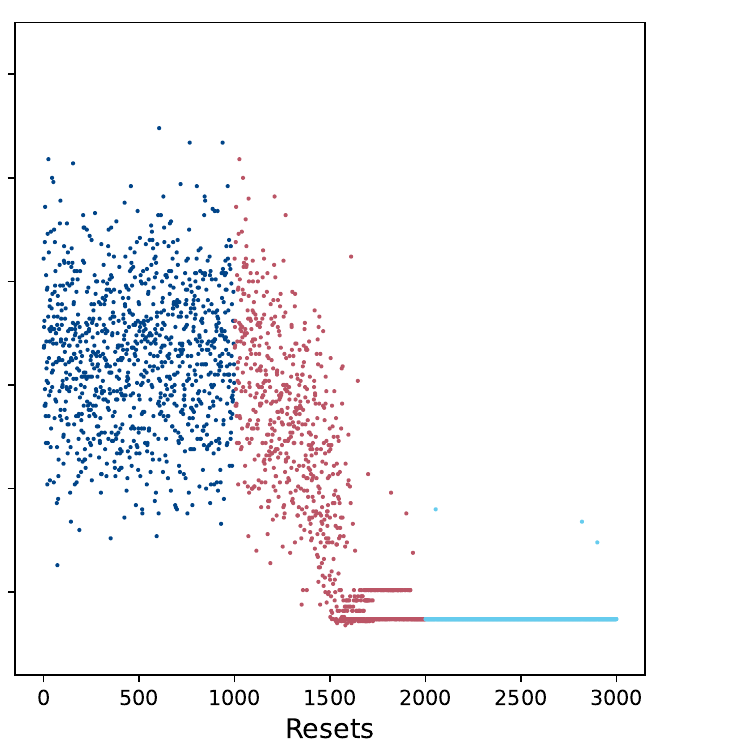}
         \caption{$\alpha=5\mathrm{e}{-7}$}
         \label{fig:N_A}
     \end{subfigure}
        \caption{\Hl[r1]{Four regimes} of learning outcomes. (a) When $\alpha$ is too low, the outcome is neither novel or appropriate. 
        \Hl[r2]{It's not novel because the distribution remains unchanged over time, nor appropriate because it hasn't converge.}
        (b) When $\alpha$ is too high, the outcome is novel \Hl[r2]{(converging to a previously unvisited state)}, but not appropriate \Hl[r2]{as that state's energy is higher than before learning}. (c) When $\alpha$ is intermediate, the typical case is an appropriate outcome, but not novel, and finally (d) for some $\alpha$, the outcome is both novel and appropriate, i.e., creative. In each plot, the points represent the energy at the end of convergence (e.g., the energies at step=1000 in Fig.~\ref{fig:E_conv}), for a set \Hl[r1]{before} learning (resets 1--1000, \Hl[r1]{dark blue}), during learning (1001--2000, \Hl[r1]{red}), and after learning (2001-3000, \Hl[r1]{light blue}).}
        \label{fig:Four-modes-of-outcome}
\end{figure}

To look at the attractor states, we will shift the perspective from the energy of the state under the update rule (Eq.~\ref{eq:state_update}) for each time step, as earlier illustrated in Fig.~\ref{fig:E_conv}, to the energy at the end of convergence \Hl[r2]{after} each reset, as depicted in Fig.~\ref{fig:Four-modes-of-outcome}. The dots in each vertical line here represent the final (attractor) energy after starting from a different initial state. This allows us to examine the distribution of the attractor energies, both with and without the influence of learning. 

In Figure \ref{fig:Four-modes-of-outcome}, these distributions are shown for four different learning parameters and 3000 resets each. Moreover, they are split into three regimes: 1000 resets \Hl[r2]{before any} learning, followed by 1000 resets with learning, followed by 1000 resets \Hl[r2]{after} learning. The first set of 1000 resets demonstrates the initial distribution of attractor states according to the randomly chosen weight matrix (Fig.~\ref{fig:W-init}). The second set demonstrates the effect of learning, and the third set demonstrates that, after learning, the system self-organizes to the learned state (except Fig.~\ref{fig:NN_NA}, which will be discussed in the next section). 

As mentioned in Sec.~\ref{sec:The-Self-Optimization-model}, the outcome of the learning process in the \gls{so} model substantially depends on the learning parameters. From Figures \ref{fig:Four-modes-of-outcome} 
, we observe four different regimes of learning outcomes depending on the learning rates for the same amount of resets (in Sec.~\ref{sec:Learning-effort} we discuss the dynamic relationship between resets and learning rates). 

For a very low learning rate $\alpha$, there is insufficient learning and there is no convergence (Fig.~\ref{fig:NN_NA}), so we cannot classify the product as either novel or appropriate. For a very high learning rate, the weight matrix is updated according to some initial state prior to convergence to a local minimum. This may result in a novel state, but it is not appropriate since it does not have any trace of the constraints of the original constraint optimization problem (Fig.~\ref{fig:N_NA_w}; see \cite{weber_use_2023-1} for a concrete example of what ``breaking constraints'' entails), resulting in higher, rather than lower, energy (Fig.~\ref{fig:N_NA}). If the learning rate is in an intermediate range, there are two possible outcomes. The typical case is convergence to a lower energy, but that energy is still within the range of the distribution \Hl[r1]{before} learning (Fig.~\ref{fig:NN_A}), so it is appropriate, but not novel.
For some $\alpha$, the converged energy of the system is below the non-learning distribution (Fig.~\ref{fig:N_A}), and hence is appropriate and novel. According the the product definition (Sec.~\ref{subsec:creativity-product}), we can consider these outcomes \emph{creative}.

In this sense, $\alpha$ can be considered the rate at which one leverages prior experience. Decreasing the learning rate too much renders the solution no longer novel, increasing it too much, it is no longer appropriate. 

\subsection{Generative Goal and Above Chance Creativity \label{subsec:above-chance}}

Note that these four \Hl[r1]{regimes} are outcomes for a single random seed determining the initial state of the system at each reset, and the change from one mode to another in terms of $\alpha$ depends strongly on the chosen random seed. 

\begin{figure}[t!p]
     \centering
     \begin{subfigure}[b]{0.49\textwidth}
       \centering
         \includegraphics[width=\textwidth]{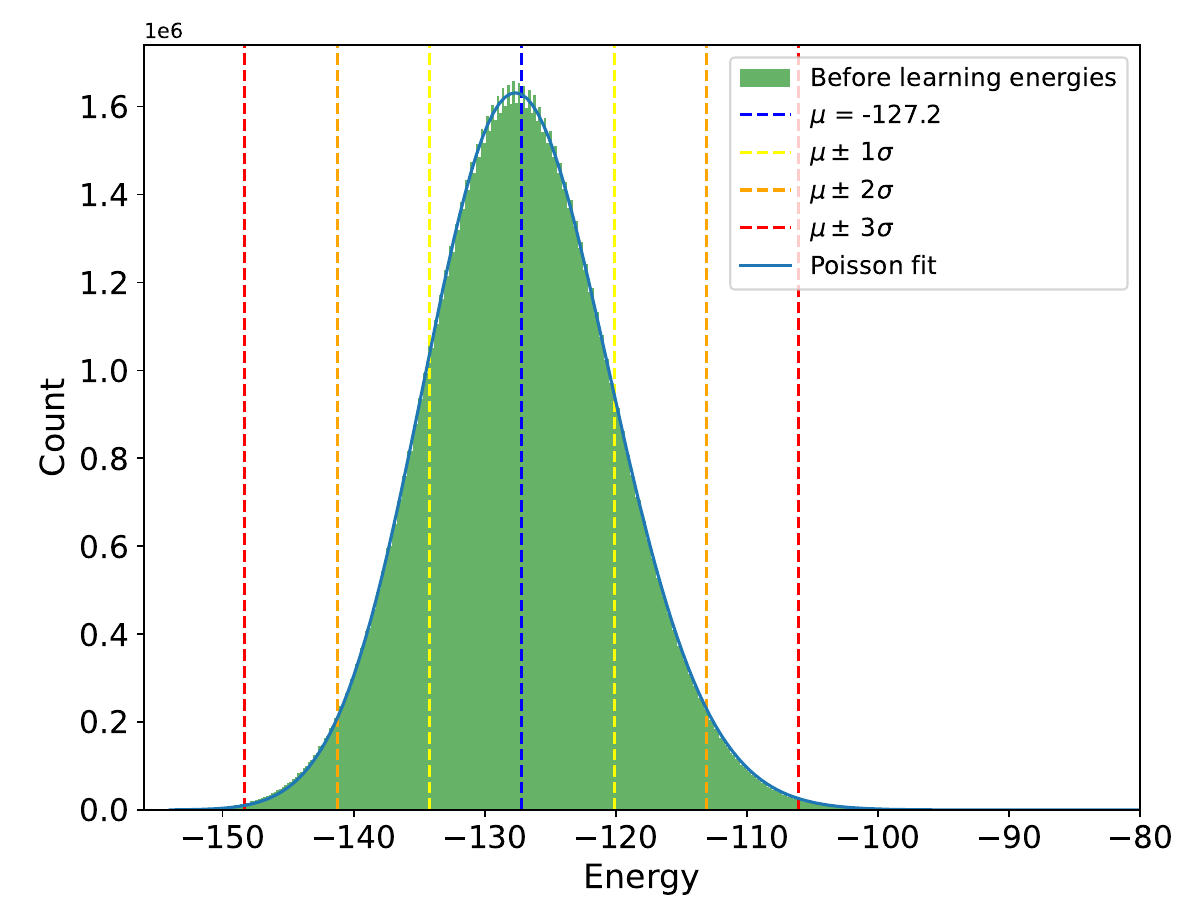}
         \caption{}
         \label{fig:nonlearning_distrib}
     \end{subfigure}
     \hfill
     \begin{subfigure}[b]{0.49\textwidth}
         \centering
         \includegraphics[width=\textwidth]{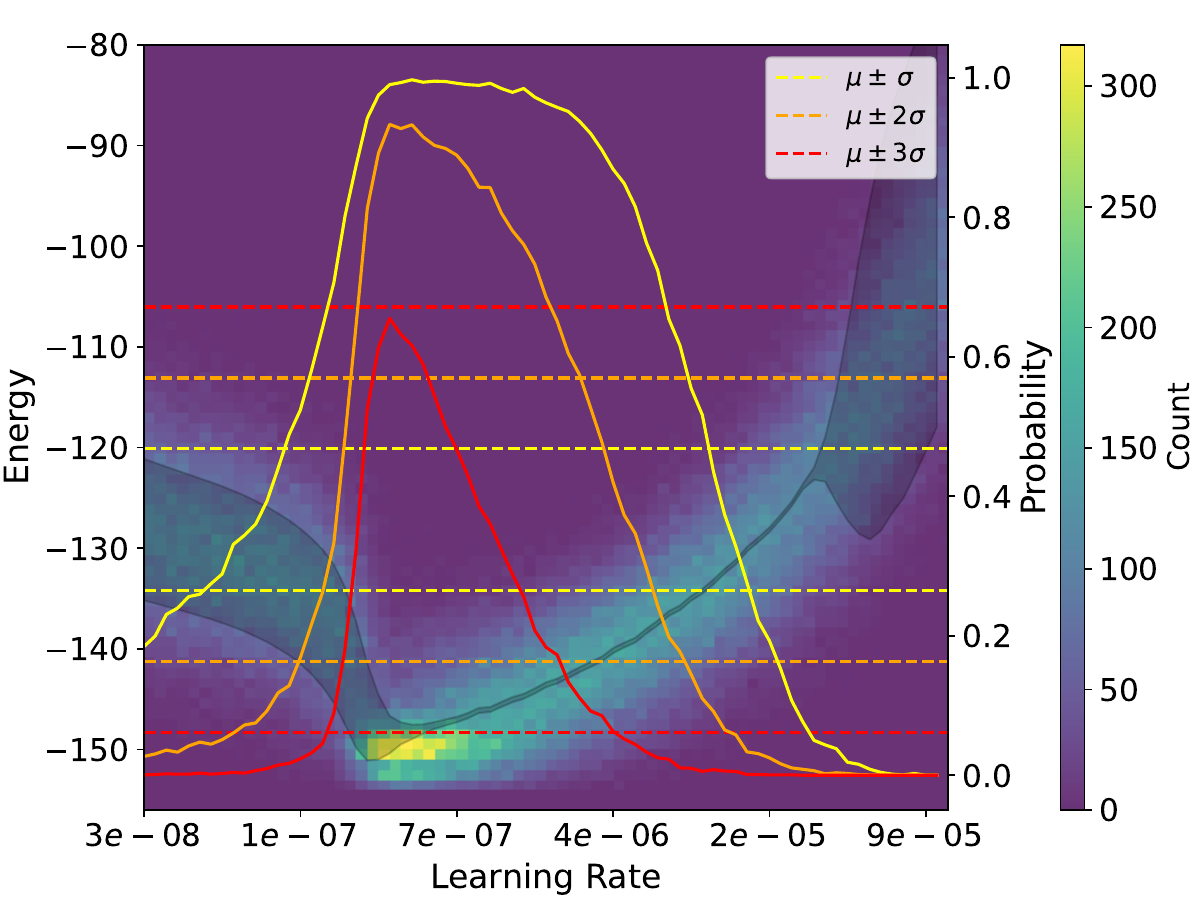}
         \caption{}
         \label{fig:learning_distrib}
     \end{subfigure}
     \begin{subfigure}[b]{0.49\textwidth}
         \centering
         \includegraphics[width=\textwidth]{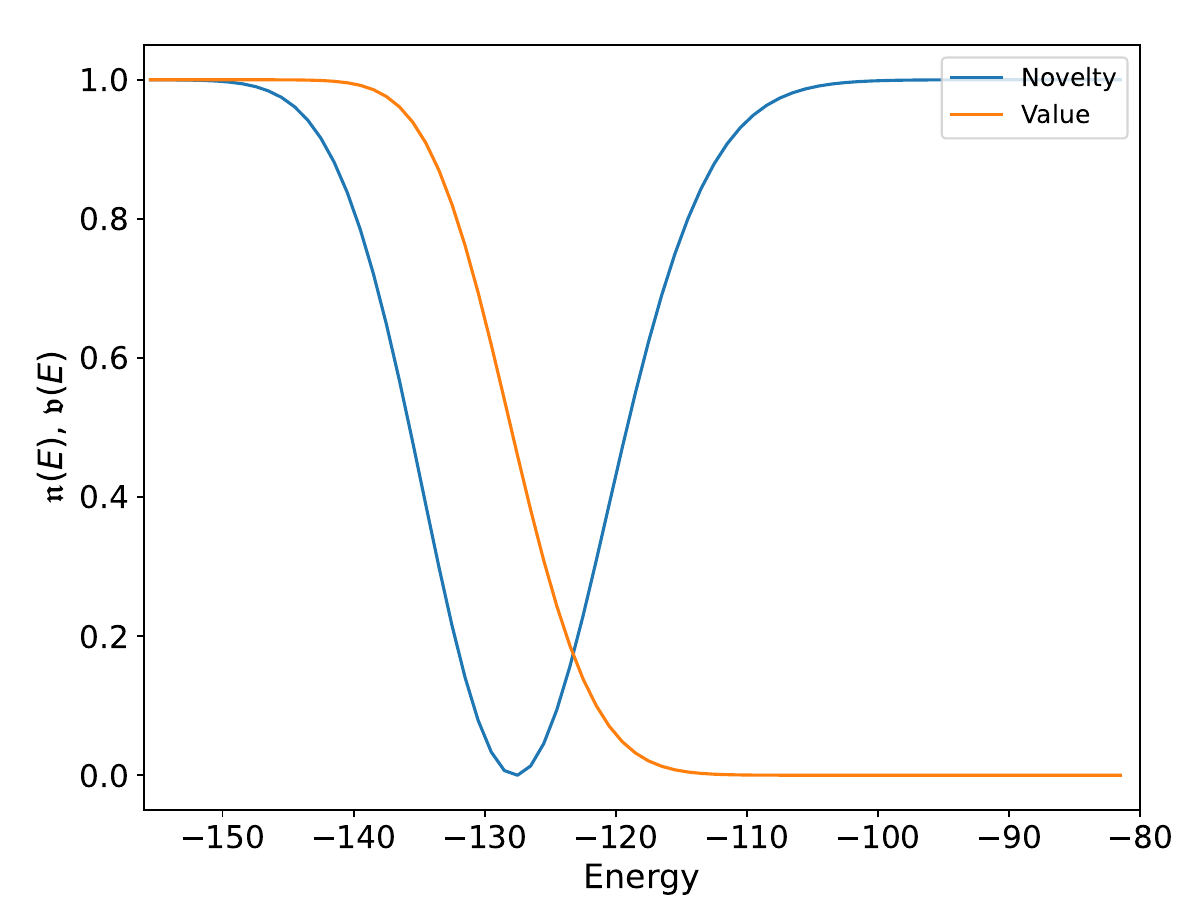}
         \caption{}
         \label{fig:novelty-value-nonlearning}
     \end{subfigure}
     \hfill
     \begin{subfigure}[b]{0.49\textwidth}
         \centering
         \includegraphics[width=\textwidth]{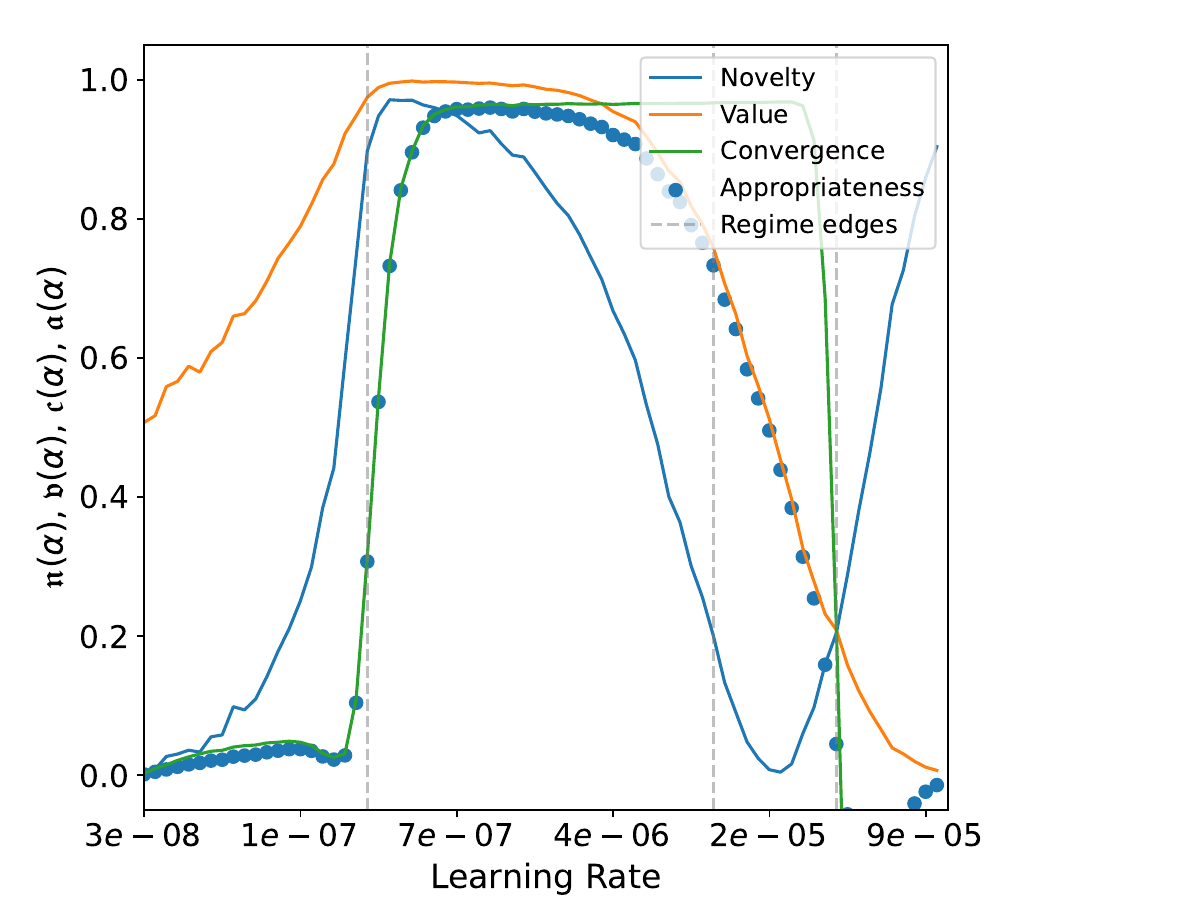}
         \caption{}
         \label{fig:novelty-value-learning}
     \end{subfigure}
        \caption{\Hl[r2]{Statistical assessment of creativity in the \gls{so} model: Before Learning (BL) vs After Learning (AL). 
        (a) Energy distribution BL, $p_{\rm BL}(E)$. The blue curve represents a Poisson fit to the data. The yellow, orange, and red dashed lines mark one, two, and three standard deviations ($\sigma_{\rm{BL}}$) from the mean,  $\mu_{\rm BL}$,
        respectively, (i.e., $\mu_{\rm BL} \pm \epsilon$). 
        (b) Energy distribution AL, $p_{\rm{AL},\alpha}(E)$, spread on a logarithmic scale. Dashed lines as in (a). 
        The corresponding solid lines show the probability $p_{\rm AL,\epsilon,\alpha}$ for the AL energy to be lower than 
        $\mu_{\rm BL} - \epsilon$.
        The black shaded area shows the seed averaged width $\sigma_{\rm{AL}}$ of the AL energy distribution indicating convergence of the system through learning. (c) Novelty $\mathfrak{n}(E)$ and value $\mathfrak{v}(E)$ as defined by Eqs.~(\ref{eq:novelty}),(\ref{eq:value}).  
        (d) Four regimes of learning outcomes (roughly indicated by dashed grey lines), from left to right: Not novel, not appropriate; novel, appropriate; not novel, appropriate; novel, not appropriate. Novelty (blue) and value (orange) are computed by integrating over all the learning outcomes for each learning rate. Appropriateness (blue dots) is the product of value and convergence (green). For a detailed mathematical description see Sec.~\ref{subsec:above-chance} and Sec.~\ref{subsec:novelty-SO}.}}
        \label{fig:novel-approp-plots}
\end{figure}

To assess whether or not the \gls{so} model pursues a generative goal, it is useful to compare whether learning can converge the system to lower energy states than are accessible to non-learning \gls{hn} at above chance probability. To this end, 
\Hl[r2]{we ran the entire simulation - 1000 resets each for before learning (BL), learning (L), and after learning (AL) stages - for 2000 seeds, for 72 learning rates. This resulted in a total of 144 million resets during BL stage. In Fig.~\ref{fig:nonlearning_distrib} we plot the probability distribution of BL final energies, $p_{\rm BL}(E)$.
The energies assume discrete values and are bounded from below justifying a Poisson fit to quantify the comparison with the learned results.}
%
\Hl[r2]{The resulting distribution has a mean $\mu_{\rm BL}=-127.2$ and a standard deviation $\sigma_{\rm BL}=7.0$. Let $\epsilon \in \{\sigma_{\rm BL}, 2\sigma_{\rm BL}, 3\sigma_{\rm BL}\}$ represent the offset from the mean  $\mu_{\rm BL}$. Thus, points at $\mu_{\rm BL} \pm \epsilon$ correspond to one, two, or three standard deviations from the mean of the BL final energy. Hereafter, we will use $\epsilon$ to denote this deviation. In Fig.~\ref{fig:nonlearning_distrib} these deviation from the mean are indicated by vertical, dashed lines. Figure~\ref{fig:learning_distrib} shows the effect of learning. The probability distribution of AL energies $p_{\rm{AL},\alpha}(E)$ is compared to the probability distribution of BL energies $p_{\rm BL}(E)$ indicated here with horizontal, dashed lines.}
We can see that the lowest energy can be reached by the system with a learning rate around $\alpha=4\mathrm{e}{-7}$.

We can then ask, how likely it is to find a lower energy state with learning than \Hl[r1]{before} learning? \Hl[r2]{To do so we first compute the probability that the AL energy is at least $\epsilon$ lower than $\mu_{\rm BL}$:} 
\begin{equation}
p_{\rm{AL},\epsilon,\alpha}=\int_E {\rm d} E \,p_{\rm{AL},\alpha}(E)(1-\Theta(\mu_{\rm BL}-\epsilon))\,,
\label{prob_NL2}
\end{equation}

\Hl[r2]{where $p_{\rm{AL},\alpha}$ is the probability for a fixed learning rate $\alpha$ to arrive at certain energy $E$, and $\epsilon \in \{\sigma_{\rm BL}, 2\sigma_{\rm BL}, 3\sigma_{\rm BL}\}$. These probabilities are shown in solid lines in Fig.~\ref{fig:learning_distrib}.}
With learning, at the optimal learning rate, $\alpha=4\mathrm{e}{-7}$, the probability that the learned energy is lower than 99.9\% ($3\sigma_{\rm BL}$) of the non-learning energies is about 65.4\% (yellow solid curve), the probability that the learned energy is lower than 98.2\% ($2\sigma_{\rm BL}$) of the non-learning energies is about 93.7\% (orange solid curve), and the probability that the learned energy is lower than 84.5\% ($1\sigma_{\rm BL}$) of the non-learning energies is about 99.6\% (red solid curve) based on the assumed Poisson distribution shown in Fig.~\ref{fig:nonlearning_distrib}. The latter shows that for a fairly broad range of learning rates (more than two orders of magnitude), the system will find a final state that is at least one sigma below the mean final energy at more than 65\% probability. 


The corresponding probabilities of finding the lowest energy state within the non-learning stage are about 0.1\%, 1.8\%, and 15.5\%, below $3\sigma_{\rm BL}$, $2\sigma_{\rm BL}$, and $1\sigma_{\rm BL}$, respectively. This shows that with learning not only can the system find low energy solutions that are less likely to be found without learning (complying with the generative goal requirements), but that these solutions are found with notably above chance probability.

\subsection{On Novelty and Appropriateness in the SO Model\label{subsec:novelty-SO}}

\Hl[r2]{As mentioned in Sec.~\ref{sec:SO-and-CC}, we use the distribution of before learning (BL) attractor states (Fig.~\ref{fig:nonlearning_distrib}) in the definition of creative products in the \gls{so} model. We quantify \textbf{novelty} as how far the final states AL are outside the distribution of BL attractor states (Fig.~\ref{fig:novelty-value-nonlearning}; blue curve), i.e.,}
\begin{equation}
\mathfrak{n}(E)=1-\frac{1}{\eta}{\rm Pois}(k(E))=1-\frac{1}{\eta}\frac{\lambda^{k(E)}}{k(E)!}{\rm e}^{-\lambda},
\label{eq:novelty}
\end{equation}
\Hl[r2]{where $\lambda=\sigma_{\rm BL}^2$, $k(E)=E-\mu_{\rm BL}+\sigma_{\rm BL}$ and $\eta={\rm Pois}(k(\mu_{\rm BL}))$. 
\textbf{Value} is quantified as how much the final state is lower than the mean of BL attractor states, $\mu_{\rm BL}$, (Fig.~\ref{fig:novelty-value-nonlearning}; orange curve), i.e., }
\begin{equation}
\mathfrak{v}(E)=1-\frac{\Gamma(\lfloor k(E)+1\rfloor,\lambda)}{\lfloor k(E)\rfloor !}
\label{eq:value},
\end{equation}
\Hl[r2]{where $\Gamma(\cdot,\cdot)$ is the upper incomplete gamma function and $\lfloor \cdot\rfloor$ is the floor function.
As a consequence, the center of the BL distribution corresponds to zero novelty and $0.5$ value and novelty can be high for both highly valuable states as well as for states that have no value at all. Another visualization of the same definition is shown in Fig.~\ref{fig:pareto} in Sec.~\ref{sec:appendix}. Having a continuous quantifier of novelty and value in this manner may be viewed as an automatic assessment of products akin to a subjective scoring method \citep{silvia_assessing_2008,beaty_semantic_2022}, where the final products can be ordered on a categorical scale from $1=$~\emph{not at all creative} (when $ \rm{novelty}=0\,AND\,\rm{value}=0$) to $5=$~\emph{very creative} (when $\rm{novelty}=1\,AND\,\rm{value}=1$). Using novelty and value in Fig.~\ref{fig:novelty-value-nonlearning} we can average over the range of learning outcomes of the $N_s=2000$ seeds and $N_r=1000$ resets for each learning rate in Fig.~\ref{fig:learning_distrib}, resulting in novelty $\mathfrak{n}(\alpha)$ and value $\mathfrak{v}(\alpha)$ after learning for each learning rate (Fig.~\ref{fig:novelty-value-learning}; blue and orange curves) defined by}
\begin{align}
\mathfrak{n}(\alpha)=\frac{1}{N_s N_r}\sum_{s,r} \mathfrak{n}(E_{s,r}), \label{eq:novelty_AL} \\
 \mathfrak{v}(\alpha)=\frac{1}{N_s N_r}\sum_{s,r} \mathfrak{v}(E_{s,r}),
\label{eq:value_AL}
\end{align}
\Hl[r2]{where $E_{s,r}$ is the single seed, single reset final energy.

The second criterion for appropriateness is convergence of the system to a single attractor state (Fig.~\ref{fig:learning_distrib}; black shaded area). \textbf{Convergence} is computed by averaging the single seed $\sigma_{{\rm AL},s}$ over all seeds $N_s$ and normalizing with respect to $\sigma_{\rm BL}$, i.e.,}
\begin{equation}
\mathfrak{c}(\alpha)=1-\sigma_{\rm AL}/\sigma_{\rm BL},
\label{eq:convergence}
\end{equation}
\Hl[r2]{where $\sigma_{\rm AL}=\frac{1}{N_s}\sum_s\sigma_{{\rm AL},s}$.
\textbf{Appropriateness} is then computed as the product of value and convergence (Fig.~\ref{fig:learning_distrib}; green curve), i.e.,} 
\begin{equation}
\mathfrak{a}(\alpha)=\mathfrak{n}(\alpha)\mathfrak{c}(\alpha).
\label{eq:approp}
\end{equation}
\Hl[r2]{Figure~\ref{fig:novelty-value-learning} shows that for low learning rates the novelty is low, because the system is very likely to reach same attractor states as without learning, and the average value of products is about 0.5, which corresponds to the mean outcome (attractor state with energy $E=-127.2$) within the non-learning distribution for (Fig.~\ref{fig:novelty-value-nonlearning}; orange curve). Appropriateness is zero for low learning rates because the system did not learn enough to converge to a single product. As learning rate increases, so are novelty and value reaching one around the optimal learning rate. For a certain range of learning rates, the products are highly novel and appropriate (i.e., creative), but as learning rate continues to increases, at some point the system starts to converge to energy states above the mean of the non-learning distribution, and accordingly their value decreases. From high learning rates, the constraints of the original problem start to break and so the novelty of the products starts to increase again. Grey dashed lines outline roughly the transition between the four different regimes as depicted in Fig.~\ref{fig:Four-modes-of-outcome}.}

According to the process perspective on creativity (Sec.~\ref{subsec:creativity-process}), the two regimes where the final product is novel \Hl[r2]{(Figs.~\ref{fig:N_A} and \ref{fig:N_NA}) }are also the regimes that can be considered generative (Sec.~\ref{subsubsec:generative-goal}). In other words, only if exhibiting novelty within these two learning rate bands can the \gls{so} model be considered an instantiation of the creative process, irrespective of arriving at an appropriate outcome. This finding complements the comparison of the process definition and the \gls{so} model with a dynamic study of the process in action and complies with \citet{green_process_2023} who state that, for a process to be considered creative, it is not necessary to arrive at a creative product.

However, novelty in the \gls{so} model is also present throughout the process due to learning changing the optimization landscape. In this sense, even if learning did not yield a novel \emph{product}, the system finds a novel \emph{path} every time it converges to a solution. In other words, even if it does not conclude what \citet{jennings_understanding_2011} denote \textit{place} search in studying creativity, it may still exhibit what they call \textit{path} search.

Our approach to categorizing regimes as appropriate or inappropriate is intentionally designed to ensure broad applicability. Unless the system could not find a solution within the provided period of time (Fig.~\ref{fig:NN_NA}) or converged on a state that disregards the original problem entirely (Fig.~\ref{fig:N_NA}), everything else 
\Hl[r2]{may be considered appropriate or novel to a certain degree.}
Figure~\ref{fig:N_NA} shows only the extreme case where almost no traces of the original constraints are left (Fig.~\ref{fig:N_NA_w}), but constraints can be modified and broken to various degrees. Moreover, some problems might not have solutions that satisfy all constraints, but we might still want to ask what is the optimal solution under the given restrictions \citep{weber_use_2023-1}. One answer to this can be found by investigating how breaking of constraints can help in solving the rest of the problem (see Sec.~\ref{sec:Discussion}).

To summarize this section, we showed that based on just one parameter, the learning rate $\alpha$, the system can go through four different regimes, where a creative outcome (both novel and appropriate) is one possibility, but various forms of non-creative outcomes also exist. The inconclusive case shown in Fig.~\ref{fig:NN_NA} is further investigated and presented in Sec.~\ref{sec:Learning-effort}.

\section{Learning Effort Pays Off\label{sec:Learning-effort}}

\begin{figure}[t!]
    \centering
    \includegraphics[width=.9\textwidth]{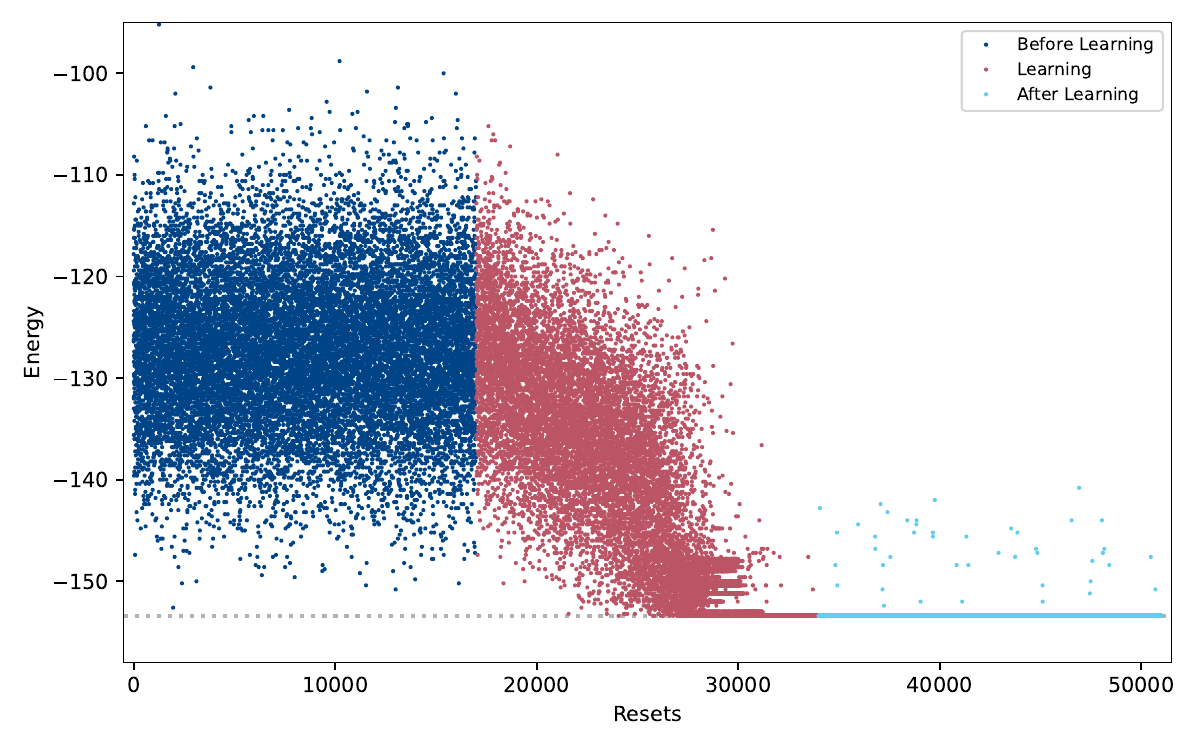}
    \caption{\label{fig:Increased-learning-effort}Increased learning effort eventually pays off. 
    \Hl[r2]{ The system can find novel solutions, even with a small learning rate of $\alpha=3\mathrm{e}{-8}$, as in Fig.~\ref{fig:NN_NA}, if more resets are used.}
    In the plot, the points represent the energy at the end of convergence, for a set \Hl[r2]{before} learning (resets 1--\Hl[r2]{17000, dark blue}), during learning (\Hl[r2]{17001--34000, red}), and after learning (\Hl[r2]{34001-51000, light blue}). Dashed line indicates the converged energy after learning.}
\end{figure}

\Hl[r2]{In Sec.~\ref{sec:SO-and-CC}, we noted that very low $\alpha$ does not yield appropriate products.}
This is, however, at least partially a result of the computational restrictions imposed on the learning.
For all the experiments presented in Fig.~\ref{fig:Four-modes-of-outcome} we used 1000 resets. This choice of the number of resets depends on the size $N$ of the system, and in Fig.~\ref{fig:N_A} it was shown that 1000 resets can be sufficient for $N=100$ for some $\alpha$ to find a creative (novel and appropriate) outcome. However, if we were to use more resets, the system would converge to a creative outcome (Fig.~\ref{fig:Increased-learning-effort}) even in the case of the very low $\alpha$ that was used in Fig.~\ref{fig:NN_NA}.
We can view the number of resets as an effort that needs to be expended to resolve the constraint problem.
The time available for completion of a task is a limited resource. Hence, the use of more resets can be thought of as an investment. From Fig.~\ref{fig:Increased-learning-effort}, we observe that as long as we are willing to put in the effort (``invest'' in more resets), even with small $\alpha$, the system can reach creative solutions. This can be viewed as a trade-off relationship.\footnote{\Hl[r1]{This trade-off is somewhat reminiscent of amortized analysis, where initial costs may lead to benefits. It is important to note that here we address the agent's resource allocation rather than providing a formal algorithmic complexity analysis.}} Small $\alpha$ requires more resets to counterbalance the slow learning dynamics and explore more of the state space. Conversely, fewer resets might suffice when $\alpha$ is well-tuned. However, there is a limit to this trade-off. If $\alpha$ exceeds an a priori unknown certain threshold, it disrupts the original problem too much, for which no amount of resets can compensate, as shown in \Hl[r2]{Fig.~\ref{fig:N_NA}}. We discuss this trade-off relationship further in Sec.~\ref{sec:Discussion}.

\section{Discussion\label{sec:Discussion}}

In this work we studied the effect of two hyperparameters of the \gls{so} model on creativity: learning rate $\alpha$ (Sec.~\ref{sec:SO-and-CC}) and the number of resets (as learning effort; Sec.~\ref{sec:Learning-effort}). 

We demonstrated that, based on modification of $\alpha$ alone, the \gls{so} model goes through \Hl[r1]{four} different regimes \Hl[r1]{(novel and appropriate, not novel but appropriate, novel but not appropriate, and neither novel nor appropriate)} of possible outcomes, of which some within a \Hl[r1]{certain} parameter range, qualify as creative products. While the corresponding simulation and particular range of the learning rate depend on the chosen seed of the source of randomness, in Sec.~\ref{subsec:above-chance} we showed that generation of these products is statistically above chance compared to the regular classical \glspl{hn}, even if we look at the least restrictive case of just $1\sigma_{\rm BL}$ below the mean final energy (65.4\% probability with learning compared to just 15.5\% without learning). 

Furthermore, we showed that there is a trade-off between $\alpha$ and the number of resets in the dynamics of the system. A smaller $\alpha$ requires more resets to offset slow learning, while an optimal $\alpha$ reduces the need for resets. A possible implication of this observation is that, given limited resources (e.g., time), the choice of the learning parameter $\alpha$ is critical. \Hl[r1]{To state the obvious: if} a process manages to resolve some problems faster than another process, it can resolve more problems than the other process in a given period of time. In the simple setup used in this work, the rate $\alpha$ is fixed by the investigator. In a real system the learning process should be able to judiciously choose $\alpha$ such as to minimize resource expenditure. One possible starting point for this could be to introduce metaplasticity in the algorithm \citep{belhadi_biologically-inspired_2023}. While we used an instantaneous reset of the entire system, it is not only possible to reset the system partially, but it may also be more temporally efficient \citep{froese_autopoiesis_2023}. Partial system resets represent a more realistic scenario,  given that in real life a reset to the entire system is akin to being struck by a lightening (the survival odds are low). Consequently, future research should investigate the impact of various reset types on the resulting creative outcomes.

In this article, we utilized the modular weight matrix (Eq.~\ref{eq:modular_W}) as an abstract problem that is hard to solve and, in its modular appearance, biologically realistic. This choice allows us to comprehensively evaluate the capabilities of the \gls{so} model, delineate the full extent of its products, and analyze its behavior with respect to learning. However, this abstract problem provided a less intuitive means to anchor appropriateness as characteristic of a creative product. It is important to note that the \gls{so} model has already demonstrated its applicability to concrete, real-world problems, as evidenced by the work of \citet{weber_use_2023-1}. To fully investigate how learning breaks constraints and how this would affect a final product's appropriateness, future research may implement the procedure outlined by \citet{weber_use_2023-1} \Hl[r2]{in a more applied setting}. 

For illustration, we used a relatively small Hopfield network of 100 nodes, but the \gls{so} model scales easily to 10,000 nodes \citep{weber_scaling_2022} \Hl[r2]{and has comparable results as depicted in Fig.~\ref{fig:Four-modes-of-outcome-10000}}. Scaling the model to thousands of nodes opens opportunities for future research into more realistic and complex problems.


Given the constantly changing landscape (i.e., state space) in the \gls{so} model, it is tantalizing to ask how it is related to \Hl[r2]{\citeauthor{boden_creativity_1998}’s} \citeyearpar{boden_creativity_1998} \Hl[r1]{influential theory} of exploratory, combinational, and transformational \Hl[r1]{creative processes} in what she calls a ``conceptual space''. \Hl[r1]{Her theory has not only shaped both the theoretical and applied discourse in computational creativity \citeg{wiggins_preliminary_2006, wiggins_searching_2006, lahikainen_creativity_2024, wiggins_framework_2019, linkola_action_2020}, but also theories of open-endedness within \gls{alife} \citep{soros_creativity_2024-1} -- another discipline that Boden contributed considerably to}.
\Hl[r1]{\textit{Combinational creativity}, according to} \citet[p.~348]{boden_creativity_1998}, \Hl[r1]{is the creation of ``novel (improbable) combinations of familiar ideas.'' \textit{Exploratory creativity} }
``involves the generation of novel ideas by the exploration of structured conceptual spaces,'' which involves a ``minimal `tweaking' of fairly superficial constraints.''
\Hl[r1]{\textit{Transformational creativity}, which is considered by many the most significant \citep{lamb_evaluating_2018},}
``involves the transformation of some (one or more) dimension of the space, so that new structures can be generated which could not have arisen before'' \citep[p.~348]{boden_creativity_1998}. 
\Hl[r1]{This distinction has been widely used in computational models \citeg{stepney_modelling_2021}, though the precise boundary between these forms of creativity remains an active area of discussion \citeg{lahikainen_creativity_2024, soros_creativity_2024-1}.}
\Hl[r1]{By these definitions, all three types of creativity can happen in the \gls{so} model. } Moreover, \citet{boden_creativity_2015} states that self-organization -- a process by which stable structures emerge spontaneously as a result of nonlinear interaction between a large number of components -- counts as a form of transformational creativity. This further supports \Hl[r2]{the} above claim, as the \gls{so} model provides a mathematical framework for a self-organized system.

\Hl[r1]{One of the most exciting insights from this study of the \gls{so} model is that the bar for creativity can be comparatively low: it becomes conceivable that the processes and products of life itself - with their mixture of complex networks, associative memory, and unstable dynamics - are creative. Definitions of creativity, thus far almost exclusively reserved to the human domain, allows for wider application and the identification of much more minimal forms of creativity all around us. This also highlights the relevance of ideas and tools developed in \gls{alife} and computational creativity to a wider range of disciplines.}

\section{Conclusion \& Future Work\label{sec:Conclusion}}

In this article, we make the case for leveraging another operational mode of \glspl{hn}, the \gls{so} model, for the study of creativity. We find that combining the relatively simple model of attractor networks with unsupervised Hebbian learning in the \gls{so} model is sufficient to constitute a creative process, and to yield creative products as solutions of the optimization process. Being able to study creativity in such networks is relevant because of their wide applicability in engineering and cognitive modeling. Attractor networks can model a wide spectrum of cognitive processes, from recognition of objects to syntax processing, navigation, planning and decision making \citep{pulvermuller_biological_2021}. Moreover and of particularly interest for this journal, \Hl[r1]{unsupervised} Hebbian learning is pertinent to all cognitive domains. 
A \Hl[r1]{minimal} mathematical model that demonstrates creativity, we may be able to both study creative processing in the brain and bring new insights into the active research area of \gls{alife}. \Hl[r1]{We highlight several opportunities for future work, from the more specific to the more general:}
\begin{itemize} 
\item 
As established in Sec.~\ref{subsubsec:attention}, we can interpret the weights of the \gls{so} model to capture various forms of attention. \citet{green_process_2023} argue that internally-directed attention is a necessary condition for creativity (while external attention is not). Consequently, modifying internal attention should affect the potential for creative outcomes. Since internal and external targets of attention as weights in the \gls{so} model can be controlled by the experimenter, future work could study this hypothesis in silico.

\item \Hl[r1]{We hold that organisms, especially in interaction with their environment,  not only face one but many, potentially overlapping challenges that could be answered through creativity. Is this continuous creative process in which goal constraints can shift captured by existing definitions, and can it be expressed through the \gls{so} model? \citeauthor{green_process_2023}'s emphasis on the importance of goal maintenance as an executive function suggests that, while constraints may evolve, creativity necessitates holding onto a set of constraints for at least some duration. This is consistent with the intuition that, while working on a creative challenge, we may notice that we actually want to solve something slightly different, changing the goal constraints as a result. However, if we never held on to the constraints for at least short intervals of time, in the extreme cases the generation process would look like a random walk - constantly switching attention to which constraints are to be satisfied without a coherent trajectory toward a solution. Given this, one question that we can ask is whether the process definition (Definition~\ref{Def:Def_Green2023}) can be used to segment longer creative processes into smaller chunks, within which goal constraints remain unchanged. Future research could delineate these stable phases in the \gls{so} model by dynamically changing the initial weights $\mathbf{W}_{0}$ that represent the agent's goal constraints and investigating how shifts in constraints contribute to or disrupt the creative process.}

\item As discussed earlier, agency and creativity not only seem both essential to life, but also appear tightly linked. In Sec.~\ref{subsec:so_agency} we outlined how agency may be represented with the \gls{so} model. This work thus paves the way for future research dedicated to investigating the dependencies between agency and creativity in more depth. Similarly, this work invites further application of creativity theory to this simple framework. 

\item Of particular interest here is to which extent such theory remains applicable, given that it has been predominantly proposed in the context of human psychology. While we follow existing work on open-endedness \citeg{soros_creativity_2024-1,taylor_evolutionary_2019,soros_is_2016} in proposing the application of Boden's theory of creative processes in Sec.~\ref{sec:Discussion}, more work is needed to understand for instance to which extent Boden's \citeyearpar{boden_creative_2003} requirement for surprise can be applied to the \gls{so} model. To this end, comparison with more recent work conceptualizing creativity as traversing a state space \citep[e.g.,][]{laroche_-sync_2024} may come in particularly handy.
\item \Hl[r1]{We have opened up our investigation by drawing parallels between the open-endedness of life and the creativity of individual organisms, to then firmly focus on the latter. Adjacent to this study of creativity in a specific formal model, our research raised more general, tantalizing questions for future work: Can research on open-endedness and creativity in \gls{alife} benefit from drawing parallels between definitions of open-ended processes and process definitions of creativity? Are definitions of open-endedness ``ontologically clean'' in the light of different perspectives on creativity and the misuse of existing creativity definitions? And, are open-endedness and creativity intersecting phenomena, or should they be better cleanly separated based on system scope and timelines to study their interaction? Crucially, these questions extend existing calls for the study of the relationship between open-endedness and creativity in that the latter have focused exclusively on the product definition of creativity and specific models (rather than a definition) of the creative process \citep{soros_creativity_2024-1}. The \gls{so} model can support this work as a specific system, popular in \gls{alife} and, as demonstrated here, capable of exhibiting creativity.}  
\end{itemize}

\Hl[r2]{Already early in evolution, learning was guiding the behaviour of living beings. In this article, we demonstrated the capacity for creativity and its connection to learning in a very minimal system and candidate for agency. We hold that it is of considerable scientific interest to understand how organisms are and can be creative in response to bodily and environmental constraints. This article supports the \gls{so} model as a fascinating candidate to study the effect of learning on creativity from the bottom up - in life as is and as it could be.}

\section{Data Availability Statement}

The code for the \gls{so} model and scripts to simulate the results presented in Figures~\ref{fig:W}–\ref{fig:Four-modes-learned-Ws} of this article are available at \Hl[r1]{an online repository} \citep{weber_self-optimization_2024}.

\section{Author Statement and Acknowledgments}

The first named author is the lead and corresponding author. We describe contributions to the paper using the CRediT taxonomy \citep{niso_contributor_2022}.
\emph{Conceptualization}: TF, NW; 
\emph{Formal analysis}: NW, CG; 
\emph{Funding acquisition}: CG; 
\emph{Methodology}: NW; 
\emph{Project administration}: TF; 
\emph{Resources}: TF, CG; 
\emph{Software}: NW; 
\emph{Supervision}: TF, CG; 
\emph{Visualization}: NW; 
\emph{Writing \-- Original Draft}: NW, CG, TF. 

Natalya Weber extends gratitude to Werner Koch for many insightful discussions and guidance on the analytical approaches of the present study, Ozan Erdem for constructive feedback regarding propositional satisfiability, and Ani Grigoryan for assistance in creating the initial illustration of the energy landscape in Fig.~\ref{fig:classical-hns-dyn}. We thank the anonymous reviewers for their detailed feedback and suggestions, which, amongst others, resulted in the additional inclusion of Figures \ref{fig:novelty-value-nonlearning}, \ref{fig:novelty-value-learning}, and \ref{fig:pareto}. This research was financially supported in part by the Aalto Science Institute (AScI) under the AScI Visiting Researcher (grant program, project number 9023608), and Helsinki Institute for Information Technology (HIIT), under the HIIT Community Support (grant number 9125064), which funded Natalya's stay at the Aalto University, Finland.

\printbibliography
\newpage
\appendix

\begin{figure}[t!]
    \centering
    \includegraphics[width=.8\textwidth]{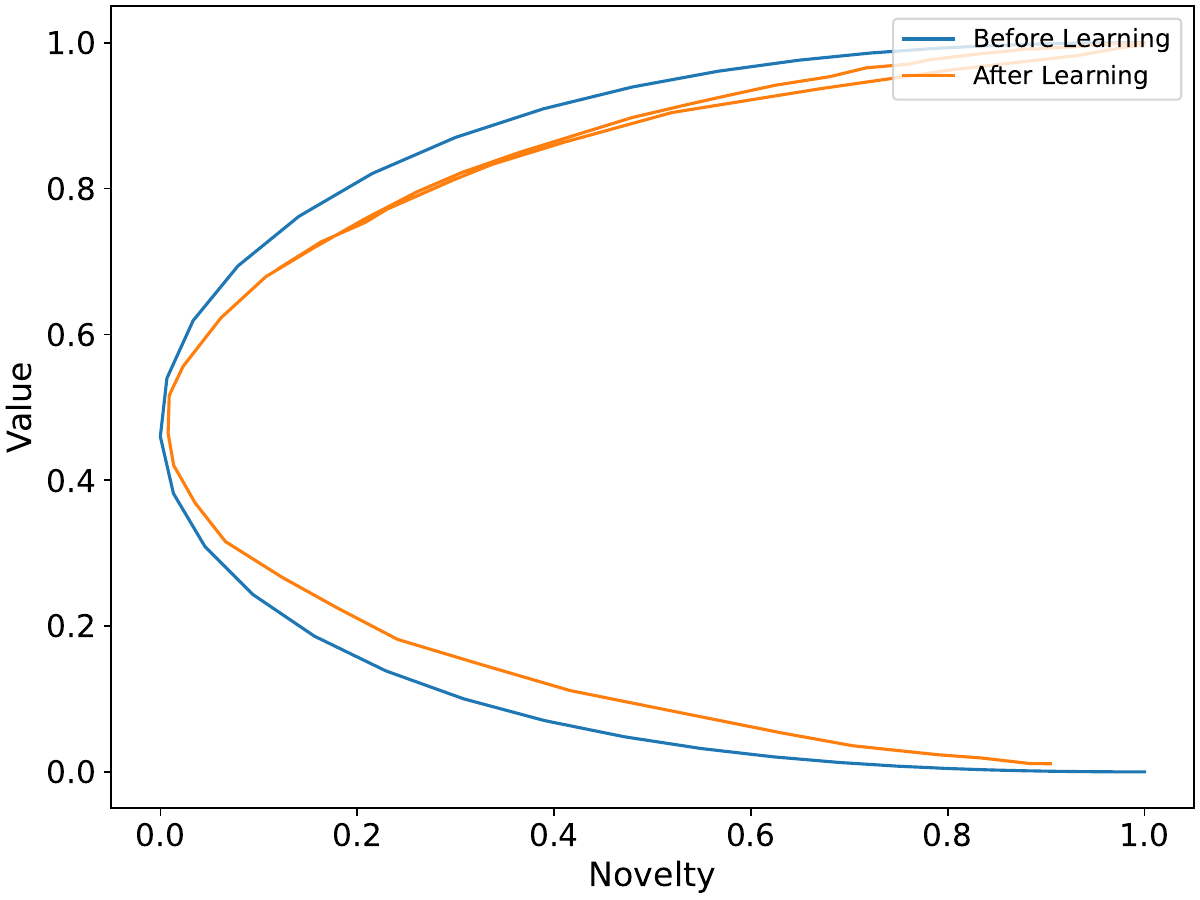}
    \caption{Parametric plots of novelty and value plotted against one another before learning (blue) and after learning (orange) using $E$ and $\alpha$  as parameter, respectively.
    }
    \label{fig:pareto}
\end{figure}

\section*{Appendix\label{sec:appendix}}

\begin{figure}[h!p]
     \centering
     \begin{subfigure}[b]{0.49\textwidth}
         \includegraphics[width=\textwidth]{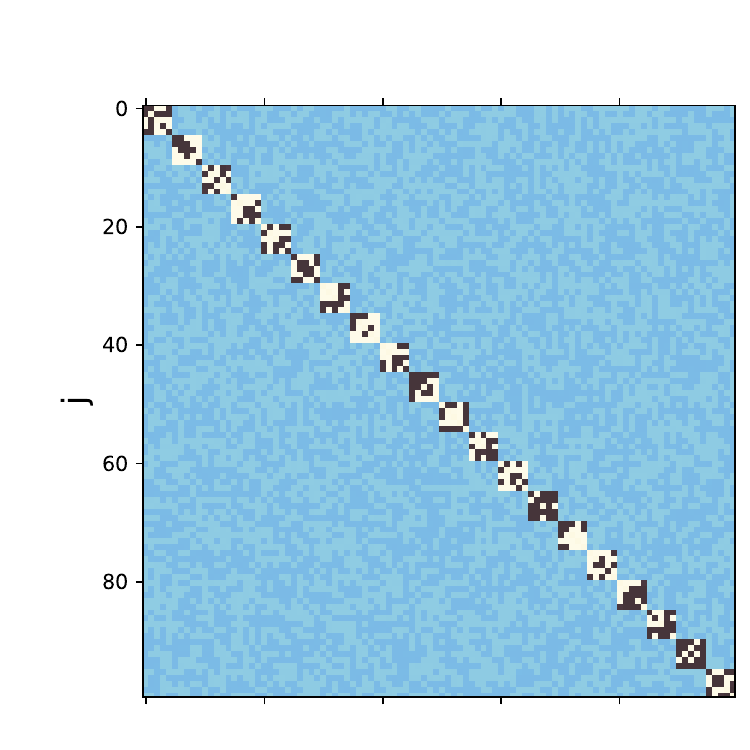}
         \caption{\Hl[r2]{$\alpha=3\mathrm{e}{-8}$}}
         \label{fig:NN_NA_w}
     \end{subfigure}
     \hfill
     \begin{subfigure}[b]{0.49\textwidth}
         \centering
         \includegraphics[width=\textwidth]{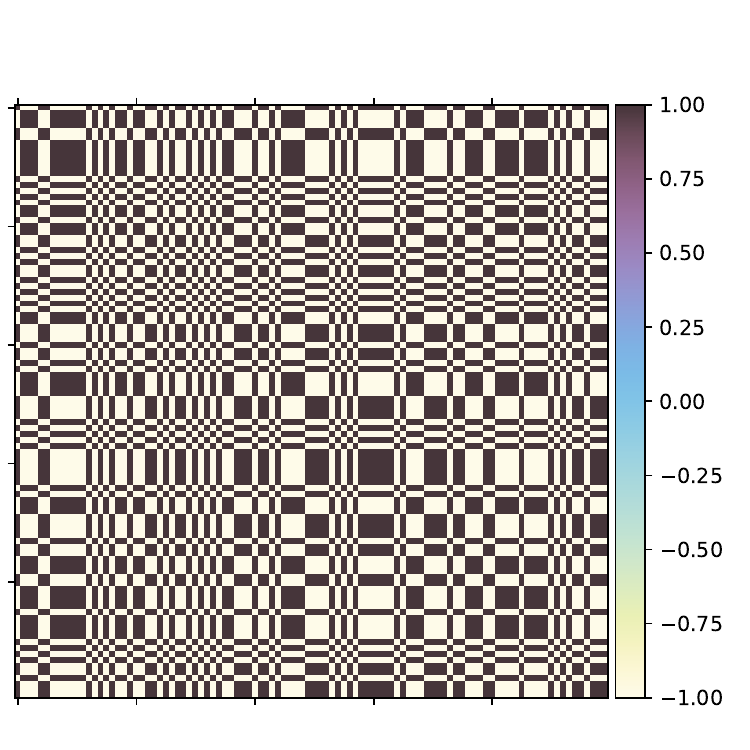}
         \caption{$\alpha=5\mathrm{e}{-5}$}
         \label{fig:N_NA_w}
     \end{subfigure}
     \begin{subfigure}[b]{0.49\textwidth}
         \includegraphics[width=\textwidth]{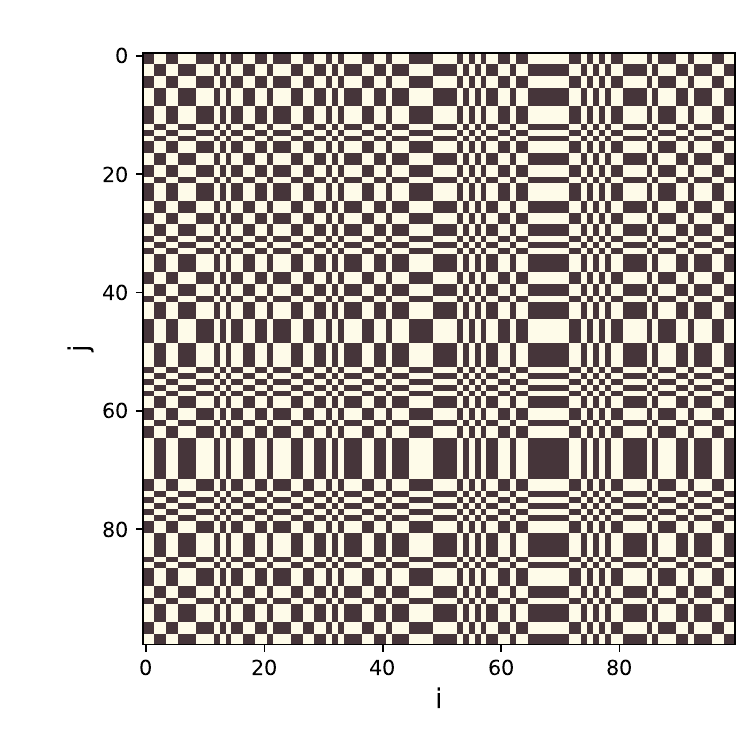}
         \caption{\Hl[r2]{$\alpha=3\mathrm{e}{-6}$}}
         \label{fig:NN_A_w}
     \end{subfigure}
     \hfill
     \begin{subfigure}[b]{0.49\textwidth}
         \centering
         \includegraphics[width=\textwidth]{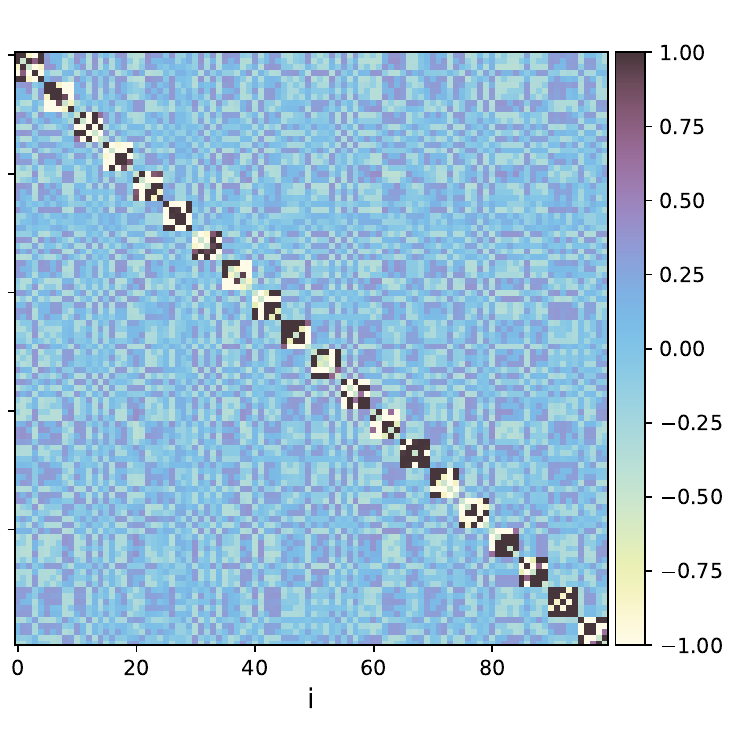}
         \caption{$\alpha=5\mathrm{e}{-7}$}
         \label{fig:N_A_w}
     \end{subfigure}
        \caption{The learned weight matrices $\mathbf{W}_{\mathrm{L}}$ for the \Hl[r1]{four different regimes} of learning outcomes presented in Fig.~\ref{fig:Four-modes-of-outcome}.}
        \label{fig:Four-modes-learned-Ws}
\end{figure}

\begin{figure}[t!p]
     \centering
     \begin{subfigure}[b]{0.49\textwidth}
       \centering
       \makebox[30pt]{}\textbf{Not Novel}
       
     \end{subfigure}
     \hfill
     \begin{subfigure}[b]{0.49\textwidth}
       \centering
       \makebox[30pt]{\textbf{Novel}}
     \end{subfigure}
     \begin{subfigure}[b]{0.49\textwidth}
        \makebox[0pt][r]{\makebox[30pt]{\raisebox{0.5\textwidth}{\rotatebox[origin=c]{90}{\textbf{Not appropriate}}}}}%
         \includegraphics[width=\textwidth]{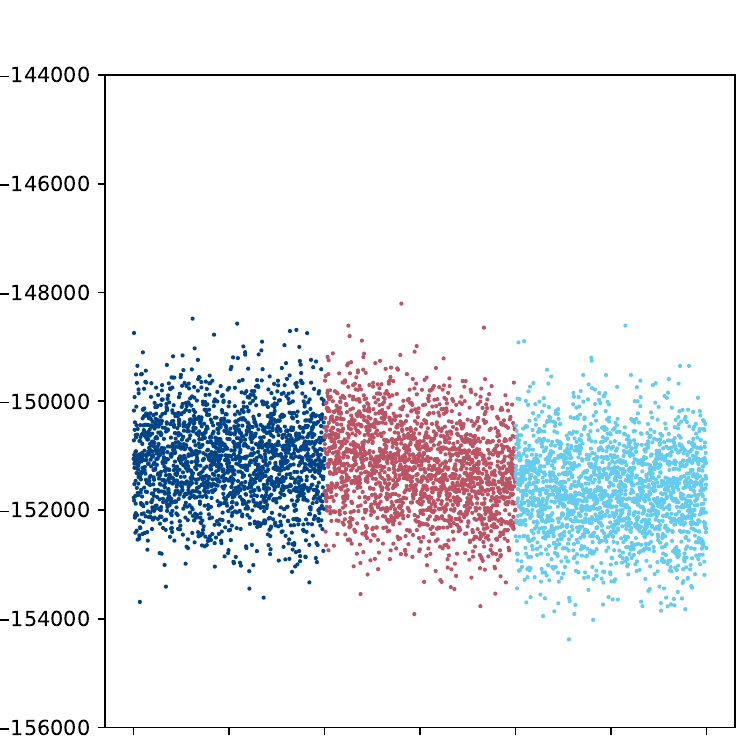}
         \caption{$\alpha=1\mathrm{e}{-10}$}
         \label{fig:NN_NA_10000}
     \end{subfigure}
     \hfill
     \begin{subfigure}[b]{0.49\textwidth}
         \centering
         \includegraphics[width=\textwidth]{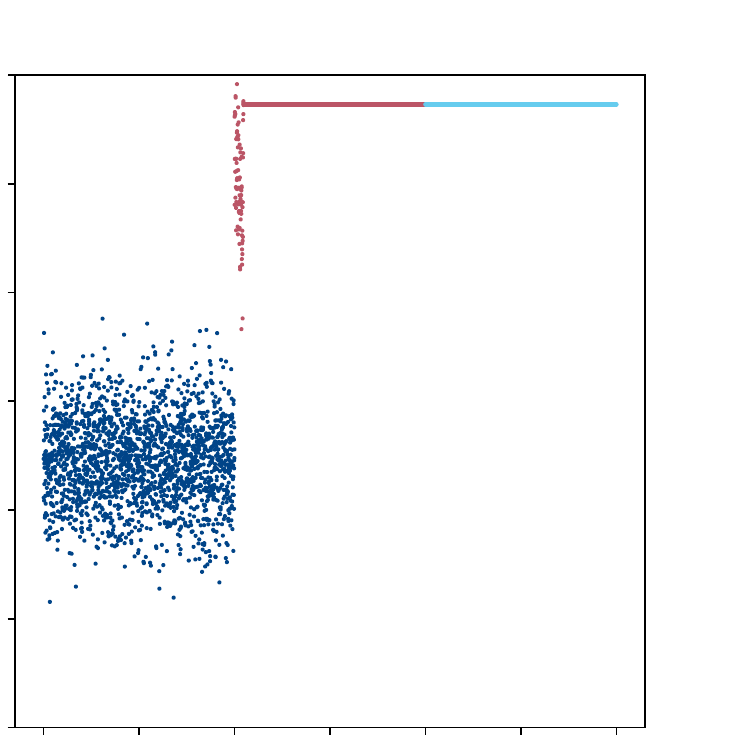}
         \caption{$\alpha=4\mathrm{e}{-9}$}
         \label{fig:N_NA_10000}
     \end{subfigure}
     \begin{subfigure}[b]{0.49\textwidth}
         \makebox[0pt][r]{\makebox[30pt]{\raisebox{0.5\textwidth}{\rotatebox[origin=c]{90}{\textbf{Appropriate}}}}}
         \includegraphics[width=\textwidth]{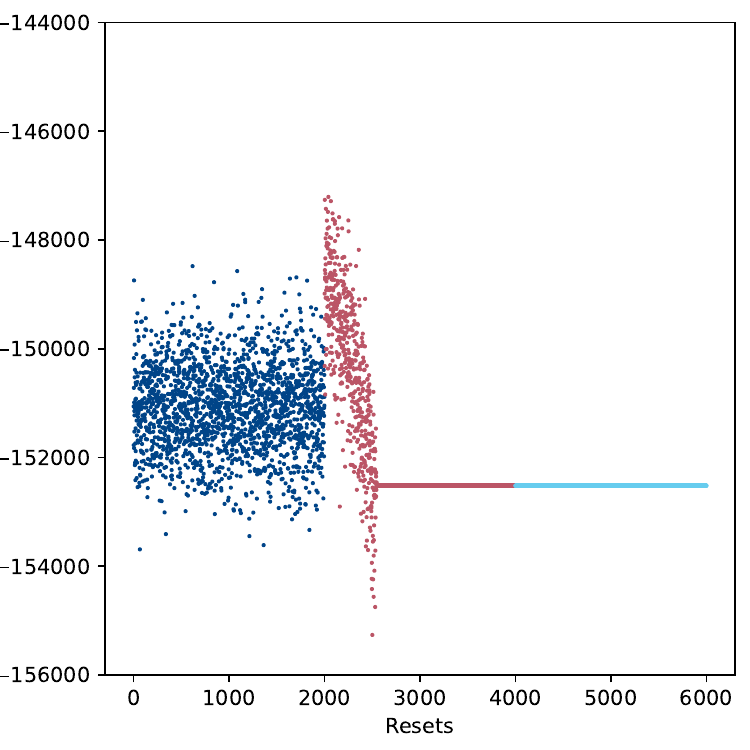}
         \caption{$\alpha=1.5\mathrm{e}{-9}$}
         \label{fig:NN_A_10000}
     \end{subfigure}
     \hfill
     \begin{subfigure}[b]{0.49\textwidth}
         \centering
         \includegraphics[width=\textwidth]{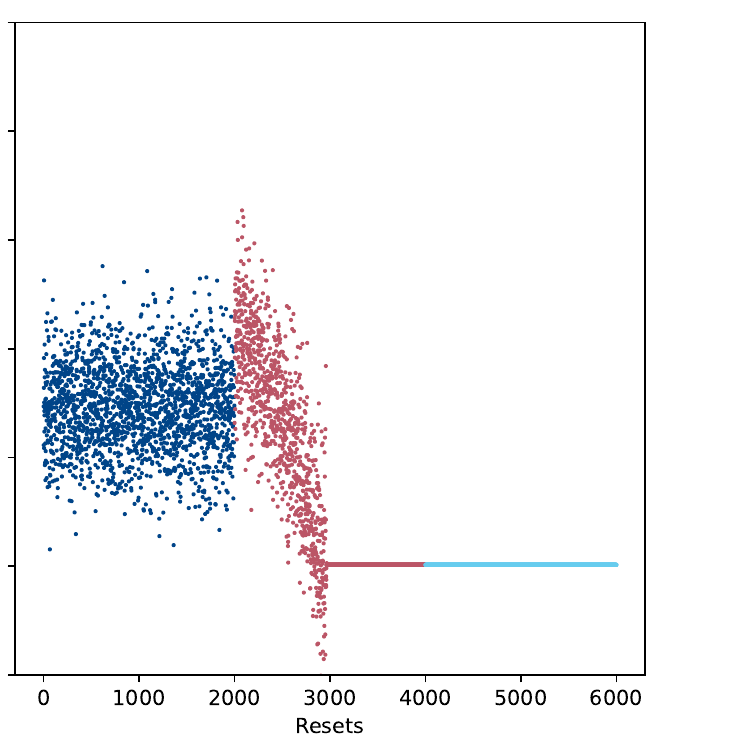}
         \caption{$\alpha=1\mathrm{e}{-9}$}
         \label{fig:N_A_10000}
     \end{subfigure}
        \caption{Four different regimes of learning outcomes for a system size of $N=10000$ nodes. Initial weights have 25 modules of size $k=400$, intra-module weights $p$ set at random to either 1 or -1, and inter-module weights set at random to either 0.1 or -0.1, and $20N$ steps were used for convergence. In each plot, the points represent the energy at the end of convergence for a set before learning (resets 1--2000, dark blue), during learning (2001--4000, red), and after learning (4001-6000, light blue).}
        \label{fig:Four-modes-of-outcome-10000}
\end{figure}

\Hl[r2]{Figure \ref{fig:pareto} shows a different way to visualize the relationship between novelty and value in the \gls{so} model as defined by Eqs.(\ref{eq:novelty}) and (\ref{eq:value}) based on the probability distribution of energies before learning (Fig.~\ref{fig:nonlearning_distrib}), and Eqs.(\ref{eq:novelty_AL}) and (\ref{eq:value_AL}) based on the probability distribution of energies after learning (Fig.~\ref{fig:learning_distrib}). The orange vs blue curves are the same curves as in Fig.~\ref{fig:novelty-value-nonlearning} and Fig.~\ref{fig:novelty-value-learning} with $E$ and $\alpha$ as the parameter, respectively.}
Figure \ref{fig:Four-modes-learned-Ws} shows the learned weights for the \Hl[r1]{four different regimes} of learning outcomes presented in Fig.~\ref{fig:Four-modes-of-outcome}. \Hl[r1]{Figure \ref{fig:Four-modes-of-outcome-10000} shows that same four different regimes of learning outcomes present in a system size of $N=10000$ nodes.}

%

\end{document}

%% file: glossary.tex
\newacronym{hn}{HN}{Hopfield Network}
\newacronym{so}{SO}{Self-Optimization}
\newacronym{cc}{CC}{Computational Creativity}
\newacronym{rnn}{RNN}{Recurrent Neural Network}
\newacronym{tsp}{TSP}{Traveling Salesperson Problem}
\newacronym[
longplural={Spurious Memories}]{sm}{SM}{Spurious Memory}
\newacronym{cam}{CAM}{Content Addressable-Memory}
\newacronym{sat}{SAT}{Propositional Satisfiability}
\newacronym{maxsat}{MaxSAT}{Maximum Satisfiability}
\newacronym{alife}{ALife}{Artificial Life}
\newacronym{eeg}{EEG}{electroencephalography}
\newacronym{ai}{AI}{Artificial Intelligence}
\newacronym{oe}{OE}{Open-Endedness}